\documentclass[10pt,twocolumn,letterpaper]{article}

\usepackage{cvpr}              % To produce the CAMERA-READY version
\usepackage{graphicx}
\usepackage{amsmath}
\usepackage{amssymb}
\usepackage{booktabs}
\usepackage{bm}
\usepackage{adjustbox}
\usepackage{float}
\usepackage[accsupp]{axessibility}

\usepackage[pagebackref,breaklinks,colorlinks]{hyperref}

\usepackage[capitalize]{cleveref}
\crefname{section}{Sec.}{Secs.}
\Crefname{section}{Section}{Sections}
\Crefname{table}{Table}{Tables}
\crefname{table}{Tab.}{Tabs.}

\def\cvprPaperID{649} % *** Enter the CVPR Paper ID here
\def\confName{CVPR}
\def\confYear{2022}

\begin{document}

\title{Learning Hierarchical Cross-Modal Association \\for Co-Speech Gesture Generation}

\author{Xian Liu$^{1}$, Qianyi Wu$^{2}$, Hang Zhou$^{1}$, Yinghao Xu$^{1}$, Rui Qian$^{1}$, Xinyi Lin$^{3}$, \\ 
Xiaowei Zhou$^{3}$, Wayne Wu$^{4}$, Bo Dai$^{5}$, Bolei Zhou$^{1}$ \\
$^1$The Chinese University of Hong Kong \quad $^2$Monash University \quad $^3$Zhejiang University\\
$^4$SenseTime Research \quad $^5$S-Lab, Nanyang Technological University \\
    {\tt\small \{alvinliu@ie, zhouhang@link, xy119@ie, qr021@ie, bzhou@ie\}.cuhk.edu.hk, }
    {\tt\small qianyi.wu@monash.edu, }\\ 
    {\tt\small \{shinylin, xwzhou\}@zju.edu.cn, }{\tt\small wuwenyan@sensetime.com, bo.dai@ntu.edu.sg}
}
\maketitle

%%%%%%%%% ABSTRACT
\begin{abstract}
Generating speech-consistent body and gesture movements is a long-standing problem in virtual avatar creation. Previous studies often synthesize pose movement in a holistic manner, where poses of all joints are generated simultaneously. Such a straightforward pipeline fails to generate fine-grained co-speech gestures. One observation is that the hierarchical semantics in speech and the hierarchical structures of human gestures can be naturally described into multiple granularities and associated together. To fully utilize the rich connections between speech audio and human gestures, we propose a novel framework named \textbf{Hierarchical Audio-to-Gesture (HA2G)} for co-speech gesture generation. In HA2G, a Hierarchical Audio Learner extracts audio representations across semantic granularities. A Hierarchical Pose Inferer subsequently renders the entire human pose gradually in a hierarchical manner. To enhance the quality of synthesized gestures, we develop a contrastive learning strategy based on audio-text alignment for better audio representations. Extensive experiments and human evaluation demonstrate that the proposed method renders realistic co-speech gestures and outperforms previous methods in a clear margin. Project page: \href{https://alvinliu0.github.io/projects/HA2G}{https://alvinliu0.github.io/projects/HA2G}.
\end{abstract}

%%%%%%%%% BODY TEXT

\section{Introduction}

When communicating with other people, we spontaneously make co-speech gestures to help convey our thoughts. Such non-verbal behaviors supplement speech information, making the content clearer and more understandable to listeners~\cite{cassell1999speech, mcneill2011hand, 2014Gesture}. Psycho-linguistic studies also suggest that virtual avatars with plausible speech gestures are more intimate and trustworthy~\cite{van1998persona}. Therefore, actuating embodied AI agents such as social robots and digital humans with expressive body movements and gestures is of great importance to facilitating human machine interaction ~\cite{salem2012generation, salem2011friendly}. To this end, researchers have explored the task of co-speech gesture synthesis~\cite{ginosar2019learning, habibie2021learning, yoon2020speech, yoon2019robots,ahuja2020style,ahuja2020no,li2021audio2gestures,qian2021speech,hasegawa2018evaluation,ferstl2020adversarial,bhattacharya2021text2gestures}, which aims at generating a sequence of human gestures given the speech audio and transcripts as input.

Traditionally, the task is tackled through building one-to-one correspondences between speech and unit gesture pairs~\cite{cassell1994animated, cassell2004beat, huang2012robot, marsella2013virtual}. Such pipelines require huge human efforts, making them inapplicable to general scenarios of unseen speech. 
Recent studies leverage deep learning to solve this problem by training a neural network to map a compact representation of audio~\cite{ahuja2020no, ginosar2019learning, habibie2021learning,li2021audio2gestures,qian2021speech} and text~\cite{ahuja2019language2pose,yoon2019robots,bhattacharya2021text2gestures,ishi2018,yoon2020speech} to holistic human pose sequence. However, such a straightforward approach fails to capture the micro-scale motions and cross-modal information, \textit{e.g.}, the subtle finger movements and the rich meanings contained in speech audio. The problem of how to learn the fine-grained cross-modal association remains unsolved.

In order to fully exploit the rich multi-modal semantics, we identify two important observations from a human gesture study~\cite{mcneill2011hand}: 
1) Different types of co-speech gestures are related to distinct levels of audio information. For example, the metaphorical gestures are strongly associated with the high-level speech semantics (\textit{e.g.}, when depicting a ravine, one would moving two outstretched hands apart and saying ``gap''), while the low-level audio features of beat and volume lead to the rhythmic gestures. 2) The dynamic patterns of different human body parts in co-speech gestures are not the same, such as the flexible fingers and relatively still upper arms. Thus it is improper to generate the upper body pose as a whole like previous studies~\cite{ginosar2019learning,habibie2021learning,yoon2020speech,li2021audio2gestures,qian2021speech,ahuja2020no,ahuja2019language2pose,ahuja2020style}.

\label{1.2}
Inspired by the discussions above,
we develop the \textbf{Hierarchical Audio-to-Gesture (HA2G)} pipeline, which generates diverse co-speech gestures. Our key insight is to build \emph{hierarchical cross-modal associations across multiple levels} between tri-modal information and generate gestures in a \emph{coarse-to-fine manner}. Specifically, two modules are devised, namely the \emph{Hierarchical Audio Learner}, and the \emph{Hierarchical Pose Inferer}. In the \emph{Hierarchical Audio Learner}, we argue that features extracted from different levels of the audio backbone capture different meanings. Additionally, text information can further strengthen the audio embedding through contrastive learning for more discriminative representations. Afterwards, based on the hypothesis that different levels of audio information contribute to different body joint movements, we associate the multi-level audio features with the hierarchical structure of human body in the \emph{Hierarchical Pose Inferer}. In particular, the association is achieved in correlation with \emph{speaking styles} encoded from speaker appearances. The hierarchy of human upper limb is predicted in a coarse-to-fine manner from shoulders to fingers like a tree structure by cascading multiple bi-directional GRU generators. In addition, we propose a novel physical regularization to enhance the realness of generated poses. Experiments demonstrate that our method synthesizes realistic and smooth co-speech gestures.

To summarize, our main contributions are three-fold: \textbf{(1)} We propose the \emph{Hierarchical Audio Learner} to extract hierarchical audio features and render discriminative representations through contrastive learning. \textbf{(2)} We propose the \emph{Hierarchical Pose Inferer} to learn associations between multi-level features and human body parts. Human poses are thus generated in a cascaded manner. \textbf{(3)} Extensive experiments show that \textbf{HA2G} can generate fine-grained co-speech gestures, which outperform state-of-the-art methods on both objective evaluations and subjective human studies.
\section{Related Work}
\noindent\textbf{Human-Centered Audio-Visual Learning.} In recent years, human-centered audio-visual learning has been extensively studied~\cite{gao20192,tian2018audio,zhou2020sep,xu2021visually,zhao2019sound,tian2020unified,tian2021cyclic,liu2022visual}. While some works utilize audio-visual correspondence to solve the problems like music-to-dance~\cite{huang2020dance, li2021dancenet3d, li2021learn}, and talking face generation\cite{zhou2019talking,chen2019hierarchical,chen2020comprises,prajwal2020lip,zhou2020makelttalk,zhou2021pose,ji2021audio-driven,liu2022semantic}, the modeling between speech and gesture remains largely unexplored. The difficulty of speech-based gesture generation lies in constructing the correspondence between speech and human gesture, which is more complicated and implicit than music-to-dance or talking face generation.

\noindent\textbf{Human Motion Synthesis.}
Synthesizing human motions has been of important interest in both computer vision and graphics, where spatial-temporal coherence of pose sequence is used to generate realistic motions~\cite{battan2021glocalnet,yan2019convolutional, zhou2020generative}. Earlier methods employ statistical models such as kernel-based probability distribution~\cite{bowden2000learning, brand2000style, galata2001learning, pullen2000animating} to synthesize human motions. Still, they fail to handle motion details, and the complicated training procedures essentially limit model capacity. Recently, the ability of deep models to generate human motions has been proven on different network architectures, where CNN-based~\cite{holden2016deep, yan2019convolutional}, RNN-based~\cite{aksan2019structured, ghosh2017learning, villegas2018neural} and GAN-based~\cite{barsoum2018hp,hernandez2019human} methods have been explored. These methods are purely visual-based with the input of history motions, while our work focuses on identifying the strong correlations between speech and gestures in conversational settings to achieve speech-driven motion synthesis.

\noindent\textbf{Audio/Text-Driven Motion Generation.}
Early works on speech-driven motion generation are mostly rule-based methods~\cite{cassell1994animated, marsella2013virtual}, where a predefined set of unit gestures and motion connecting rules are designed manually. With the development of deep learning, data-driven approaches have demonstrated superior performance. Some works map speech text information to co-speech gestures~\cite{ishi2018,yoon2019robots,bhattacharya2021text2gestures,ahuja2019language2pose}. Yoon \textit{et al.}~\cite{yoon2019robots} resort to RNN to map from utterance text to upper body gestures. Some methods use speech audio signals to drive gestures~\cite{ahuja2020style, ferstl2020adversarial, ginosar2019learning, habibie2021learning, hasegawa2018evaluation, qian2021speech, li2021audio2gestures}. For example, Ginosar \textit{et al.}~\cite{ginosar2019learning} collect a 2D speaker-specific gesture dataset and train the model with an adversarial loss. To make gestures more expressive, Habibie \textit{et al.}~\cite{habibie2021learning} lift the 2D pose to 3D and generate facial expressions simultaneously. However, all of their methods learn a model for each speaker individually, which makes it hard to transfer to general scenes and limit speaker styles to a tiny number. Besides, either audio- or text-driven motion generation methods fail to consider messages from both modalities, which motivates recent methods to jointly tackle multi-modal information~\cite{kucherenko2020gesticulator, yoon2020speech,ahuja2020no}. Specifically, Yoon \textit{et al.}~\cite{yoon2020speech} propose to encode the trimodal feature embeddings of text, audio, speaker identity and concatenate them together to pass a decoder. But they fail to fully make use of multi-level features. Further, the dynamic patterns of different human body parts are diverse when people talk, \textit{e.g.}, the range and frequency of co-speech finger and arm movement are not the same, which makes it unreasonable to learn holistic human pose directly. In this work, we propose to extract hierarchical audio features with a contrastive learning strategy to excavate cross-modal messages at multiple granularities and learn co-speech gestures in a coarse-to-fine manner.
\section{Our Approach}

\begin{figure*}[t]
    \centering
    \includegraphics[width=1\linewidth]{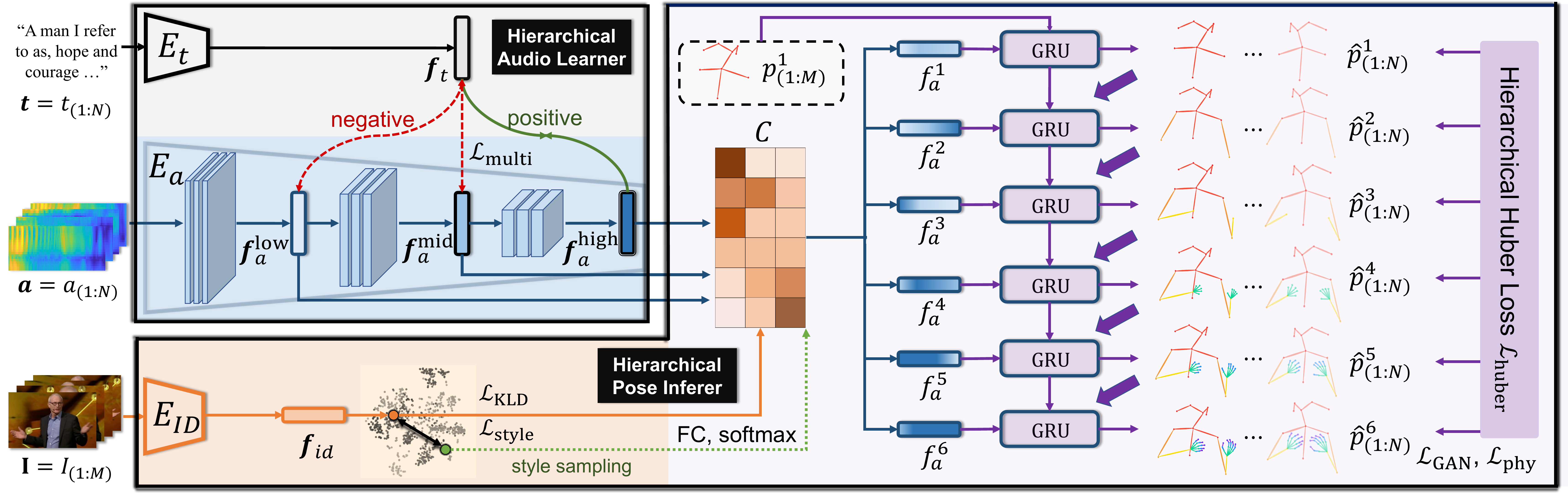}
    \caption{\textbf{Illustration of the Hierarchical Audio-to-Gesture (HA2G).} In Hierarchical Audio Learner, $E_a$ encodes speech audio $\mathbf{a}$ into multi-level audio features $\bm{f}^{\mathrm{low}}_a$, $\bm{f}^{\mathrm{mid}}_a$ and $\bm{f}^{\mathrm{high}}_a$ (\textcolor[rgb]{0.18039215686275, 0.45882352941176, 0.71372549019608}{blue}). The speech transcript $\mathbf{t}$ is encoded by $E_t$ into text features $\bm{f}_{t}$ (\textcolor[rgb]{0.37647058823529, 0.37647058823529, 0.37647058823529}{grey}). Then a contrastive learning strategy is used to enforce the discriminative audio feature extraction by attracting text feature and high-level audio feature (\textcolor[rgb]{0.32941176470588, 0.6, 0.2078431372549}{
   green}) while repelling from low/mid-level features (\textcolor[rgb]{0.83921568627451, 0, 0}{red}). In Hierarchical Pose Inferer, the reference frames $\mathbf{I}$ are encoded by $E_{\mathrm{ID}}$ to represent speaker's identity $\bm{f}_{id}$ (\textcolor[rgb]{0.92941176470588, 0.49019607843137, 0.1921568627451}{orange}), which is then transformed to style coordinator $C$ for multi-level feature blending ($\bm{f}_{a}^1$, ..., $\bm{f}_{a}^6$). Finally the co-speech gestures $\hat{\bm{p}}_{(1:N)}^6$ are generated by cascaded bi-GRU based on initial poses $\bm{p}_{(1:M)}^1$ in a coarse-to-fine manner (\textcolor[rgb]{0.43921568627451, 0.18823529411765, 0.62745098039216}{
   purple}).}
    \label{fig:framework}
\end{figure*}

We present \textbf{Hierarchical Audio-to-Gesture (HA2G)} that generates a target person's co-speech gestures given speech audio. The generated poses are conditioned on speaker identity and initial poses. Following Yoon \textit{et al.}~\cite{yoon2020speech}, text information can be provided additionally. The whole pipeline is illustrated in Fig.~\ref{fig:framework}. In this section, we first formulate the problem in Sec.~\ref{sec:3.1}, and then elaborate the \emph{Hierarchical Audio Learner} which extracts hierarchical audio features in Sec.~\ref{sec:3.2}. Sec.~\ref{sec:3.3} introduces the \emph{Hierarchical Pose Inferer} to perform multi-level feature blending and co-speech gesture synthesis. Finally, training objectives for gesture generation are described in Sec.~\ref{sec:3.4}. 

\subsection{Problem Formulation}

\label{sec:3.1}
Large amounts of speaking videos with clear co-speech gestures are used for training. Given a video with $N$ frames $\mathbf{V} =  \{I_1, \dots, I_N\}$, the skeletal poses of the upper body can be denoted as $\mathbf{p} = \{\bm{p}_1, \dots, \bm{p}_N \mid \bm{p}_i = \left[\bm{d}_{i,1}, \bm{d}_{i,2}, \dots, \bm{d}_{i,J-1}\right]\}$. Each $\bm{p}_i$ is represented as the concatenation of unit direction vectors $\bm{d}_{i,j}$ between $J$ joints. The goal of our model $G$ is to use the video's accompanying speech audio sequence $\mathbf{a} = \{\bm{a}_1, \dots, \bm{a}_N\}$ 
to recover $\mathbf{p}$ according to target's identity representation $\bm{f}_{id}$ and initial poses $\{\bm{p}_1, \dots, \bm{p}_M\}$. Following the setting of Yoon \textit{et al.}~\cite{yoon2020speech}, the text transcripts $\mathbf{t} = \{\bm{t}_1, \dots, \bm{t}_N\}$ are also provided for training. 
With encoder $E_a$ for audio information extraction, the overall objective can be written as:
\begin{equation} \label{eq:formulation}
\mathop{\arg\min}_{G, E_a} || \mathbf{p} - G ( E_a(\mathbf{a}) | \bm{f}_{id}, \bm{p}_1, \dots, \bm{p}_M) ||.
\end{equation} 

\subsection{Hierarchical Audio Learner}
\label{sec:3.2}

\noindent\textbf{Hierarchical Audio Feature Extraction.}
In most previous studies~\cite{ginosar2019learning,habibie2021learning,yoon2020speech, ahuja2020style, li2021audio2gestures, qian2021speech}, only high-level audio features are extracted to guide the synthesis of desired movements. However, it has been discussed that different semantics in audios contribute to different granularities in the movements of human poses~\cite{mcneill2011hand,loehr2012temporal}, which has been mostly ignored in previous works. We identify that such multi-level audio information could be inferred from the hierarchy of an audio encoder $E_a$  to improve the generation resolution. Notably, the rich semantics of hierarchical feature maps embedded at different layers of a deep neural network have been explored in other deep learning tasks~\cite{lin2017feature, ronneberger2015u, wang2020deep}. Therefore, the output deep feature $\bm{f}^{\mathrm{high}}_a$ of $E_a$, the feature $\bm{f}^{\mathrm{mid}}_a$ encoded in the middle of the audio encoder and the feature $\bm{f}^{\mathrm{low}}_a$ encoded in the shallow of $E_a$ are specifically leveraged.
We expect $\bm{f}^{\mathrm{low}}_a, \bm{f}^{\mathrm{mid}}_a, \bm{f}^{\mathrm{high}}_a$ to represent the low, middle and high level audio features respectively, as shown in blue block of Fig.~\ref{fig:framework}. These hierarchical features are used for inferring poses in Sec.~\ref{sec:3.3}.

\noindent\textbf{Contrastive Learning Strategy.}
Though we expect the audio features can be learned automatically given the property of the encoder, additional text can further enforce the embedding of our desired information. Transcripts, which represent high-level linguistic information, can be directly recognized by Automatic Speech Recognition (ASR) models~\cite{hinton2012deep, nefian2002dynamic, yu2020audio} from speech. Thus we propose to learn the association between provided transcripts and audios in a simple yet effective manner with contrastive learning.
Our strategy is to leverage the natural synchronization between text and audio. While the high-level audio features should reflect the temporally-aligned transcripts, text can in turn encourage mid- and low-level audio features to capture crucial speech content-irrelevant information such as tone and cadence. 

 Specifically, we denote the feature extracted by the text encoder from transcript $\textbf{t}$ as $\bm{f}_t = E_t(\mathbf{t})$. In our contrastive learning formulation, 
 the high-level audio features aligned to the transcript serve as positive examples, which are denoted as $\bm{f}^\mathrm{high}_{a+}$.
 Then we design two types of negative samples: (1) Firstly, high-level features extracted at other time steps, or from other clips are selected as negative samples to enforce the high-level audio feature capture correct semantic information from the aligned text; (2) Secondly, the low/mid-level audio features are expected to be discriminative to reflect other audio information rather than high-level semantics. Therefore, we enforce them all to repel the text feature. With the similarity function defined as $sim(f_1, f_2) = \frac{\bm{f}_1\cdot \bm{f}_2}{|\bm{f}_1||\bm{f}_2|}$, we can compute the final multi-level contrastive loss as:
 \begin{equation} 
    \label{eq:contrastive-multi}
    \mathcal{L}_\mathrm{multi} = -\log\frac{\mathrm{exp}(sim( \bm{f}_t, \bm{f}^\mathrm{high}_{a+}) /\tau)}{\sum_{i=1}^K \sum_{l \in L} \mathrm{exp}(sim( \bm{f}_t, \bm{f}^l_{a(i)})/\tau)
    },
\end{equation}
where $L=\{\mathrm{low, mid, high}\}$, and $\bm{f}^\mathrm{low}_{a(i)}$, $\bm{f}^\mathrm{mid}_{a(i)}$,  $\bm{f}^\mathrm{high}_{a(i)}$ denote the $i$-th sample of low/mid/high-level audio feature, respectively. $K$ is the number of samples and $\tau$ is the temperature parameter that controls the concentration of distribution.

\subsection{Hierarchical Pose Inferer}
\label{sec:3.3}
As discussed in Sec.~\ref{1.2}, different levels of audio features contribute to different hierarchies of human poses. Thus we propose to hierarchically infer gestures for more delicate audio-based control. To this end, we detach the joints from human body ends (fingers) to the main structure (spine) in $H$ stages as illustrated in Fig.~\ref{fig:framework} (right). However, two questions still remain: 1) How to associate multiple levels of audios with different levels of joints; 2) How to supervise coarse-to-fine generation process.

\noindent\textbf{Multi-Level Feature Blending with Style Coordinator.} Our solution to the first question is to learn automatic feature blending schemes for different levels depending on a person-related style coordinator. 
As human gestures corresponding to the same speech are diverse across persons, the idea of learning person-specific styles has been adopted in various audio-driven animation tasks~\cite{yoon2020speech,ahuja2020style}.
In this work, the style coordinator should be responsible for finding the suitable ratio among hierarchical audio features that contributes to each level of motion hierarchy.

Different from~\cite{yoon2020speech} that uses one-hot labels to represent identities, we leverage a more general form by learning from the appearances of reference frames. The encoder $E_\mathrm{ID}$ is used to extract identity feature from a few frames, $\bm{f}_{id}=E_\mathrm{ID}(I_1, \dots, I_M)$. 
Then through a linear layer and softmax function, $\bm{f}_{id}$ is transformed into the style coordinator $C\in \mathbb{R}^{3 \times H}$, where $\sum^3_{i=1} C[i, h] = 1$. In this way, we can associate multi-level audio features with hierarchical body parts by linear blending:
\begin{equation} 
    \label{eq:speakerid}    
    \bm{f}^{{h}}_a = C[1, h] \cdot \bm{f}^{\mathrm{low}}_a + C[2, h] \cdot \bm{f}^{\mathrm{mid}}_a + C[3, h] \cdot \bm{f}^{\mathrm{high}}_a,
\end{equation}
where $\bm{f}^{{h}}_a$ denotes the blended audio feature for the $h$-th motion hierarchy. The procedure is illustrated in the middle of Fig.~\ref{fig:framework}. To further facilitate style sampling at the inference stage, the Kullback–Leibler (KL) divergence loss $\mathcal{L}_{\mathrm{KLD}}$ between the feature space of $\bm{f}_{id}$ and $\mathcal{N}(0, \mathrm{I})$ is adopted to assume Gaussian style embedding distribution.

\noindent\textbf{Coarse-to-Fine Pose Generation.}
We follow the human body dynamic rules to design a $H$-level ($H$ = 6) body hierarchy (Fig.~\ref{fig:framework} right).
At each level, the generation is affected by both the inferred pose from the previous level and the current level's audio feature rendered by the style coordinator. Such an idea is also similar to previous coarse-to-fine network designs~\cite{newell2016stacked}.

In particular, we leverage the bi-directional GRU as motion decoder since the recurrent structure effectively captures spatial-temporal dependency in human motion as proved in~\cite{0Convolutional, wei2019motion}. 
With the hierarchical audio feature of the $h$-th level $\bm{f}_{a}^h = \{\bm{f}^h_{a(1)}, \dots, \bm{f}^h_{a(N)}\}$, the $h$-th level co-speech gesture $\hat{\mathbf{p}}^h = \{\hat{\bm{p}}_1^h, ..., \hat{\bm{p}}_N^h\}$ is generated by:
\begin{equation} 
    \label{eq:gru}    
    \hat{\bm{p}}_i^h = [\textbf{h}_i; \hat{\bm{p}}_i^{h-1}; \bm{f}_{a(i)}^h] * W^h + \bm{b}^h,
    \textbf{h}_i=\mathrm{GRU}(\textbf{h}_{i-1}, \hat{\bm{p}}_{i-1}^h),
\end{equation}
where $\textbf{h}_i$ is the $i$-th hidden state, $[\cdot;\cdot]$ is the concatenation operation and $*$ is the matrix multiplication. $W^h \in \mathbb{R}^{(d_s+d_p^{h-1}+d_a) \times d_p^h}$ and $\bm{b}^h \in \mathbb{R}^{d_p^h}$ are parameters where $d_s$, $d_a$ and $d_p^h$ are the dimensions of hidden state, audio feature and the $h$-th level pose $\hat{\mathbf{p}}^h$, respectively.
Note that the poses of the first $M$ frames serve as initial poses and are denoted as $\hat{\mathbf{p}}^0=\{\bm{p}^0_1, ..., \bm{p}^0_M, 0, ..., 0\}$. In this way, fine-grained correspondences between audio sequence and co-speech gestures are jointly built in a coarse-to-fine manner. The last layer's output $\hat{\mathbf{p}}^H$ from the hierarchy is our desired result. This procedure is depicted in the right part of Fig.~\ref{fig:framework}.

\subsection{Training Objectives for Gesture Generation}
\label{sec:3.4}
\noindent\textbf{Reconstruction Huber Loss.}
The generation process is constrained via a hierarchical Huber loss~\cite{huber1992robust} by measuring the distances between generated samples $\hat{\bm{p}}_i^h$ and ground truth ${\bm{p}}_i^h$:
\begin{equation} 
    \label{eq:pose}    
    \mathcal{L}_{\mathrm{huber}} = \mathbb{E}\left[\frac{1}{HN}\sum_{h=1}^H\sum_{i=1}^N\mathrm{HuberLoss}(\bm{p}_i^h, \hat{\bm{p}}_i^h)\right],
\end{equation}
where $H$ is the number of motion hierarchy and $N$ is the length of gesture sequence. We feed the blended audio feature to cascaded bi-GRU as generator $G$ and leverage an adversarial loss for preserving realism following~\cite{ginosar2019learning,yoon2020speech}:
\begin{equation} 
\begin{aligned}
    \label{eq:gan}    
    \mathcal{L}_{\mathrm{GAN}} &= \min_{G} \max_{D} \mathbb{E}_{\mathbf{p}}\left[\log D(\mathbf{p})\right] \\ &+\mathbb{E}_{\mathbf{a}}\left[\log(1- D( G (E_a(\mathbf{a}) | \bm{f}_{id}, \bm{p}_{1:M}))\right].
\end{aligned}
\end{equation}
\noindent\textbf{Style Diverging Loss.}
To further avoid posterior collapse on speaker identity $\bm{f}_{id}$, we guide the generator to synthesize different poses with diverse style input following~\cite{yoon2020speech}. Assuming that $\hat{\mathbf{p}}(\bm{f}_{id})$ is the predicted pose depending on identity feature $\bm{f}_{id}$, we have:
\begin{equation} 
    \label{eq:style}    
    \mathcal{L}_{\mathrm{style}} = -\mathbb{E}\left[\min \left(\frac{\mathrm{HuberLoss}(\hat{\mathbf{p}}(\bm{f}_{id(1)}), \hat{\mathbf{p}}(\bm{f}_{id(2)}))}{\|\bm{f}_{id(1)} - \bm{f}_{id(2)}\|_1}, \epsilon\right)\right],
\end{equation}
where $\bm{f}_{id(1)}$, $\bm{f}_{id(2)}$ are two different speaker identities and $\epsilon$ is the numerical clipping parameter. 

\noindent\textbf{Physical Constraint.}
Previous methods on co-speech gesture generation mostly fail to consider human physical constraint, which leads to unnatural poses and incoherent results. 
Therefore, we propose to add restrictions on the included angle between bones to ensure reasonable human poses. 
Concretely, the pose is represented as directional vectors, thus the angle between consecutive bone vectors must obey physical rules. We specifically calculate the mean and variance of each angle within TED-Expressive dataset, and expect our generated ones to fall within such a Gaussian distribution. The loss function for the physics constraint is the log-likelihood function:
\begin{equation} 
    \label{eq:physical}
    \mathcal{L}_{\mathrm{phy}} =-\sum_{j=1}^{J-1}\log\mathcal{N}(\theta_j; \mu_j, \sigma_j^2)
\end{equation}
where $\theta_j$ is the $j$-th bone angle value, $\mu_j$ and $\sigma_j^2$ are the mean and variance of the $j$-th angle, respectively.

The overall learning objective for the whole framework is as follows:
\begin{equation} 
\begin{aligned}
    \label{eq:losstotal}
    \mathcal{L}_{\mathrm{total}} &= \mathcal{L}_{\mathrm{GAN}} + \lambda_h \mathcal{L}_{\mathrm{huber}} + \lambda_p \mathcal{L}_{\mathrm{phy}}\\ &+ \lambda_s \mathcal{L}_{\mathrm{style}} + \lambda_k \mathcal{L}_{\mathrm{KLD}} + \lambda_c  \mathcal{L}_{\mathrm{multi}},
\end{aligned}
\end{equation}
where the $\lambda_h, \lambda_p, \lambda_s, \lambda_k,\lambda_c$ are weight coefficients. At the training stage, the hierarchical audio encoder $E_a$, text encoder $E_t$, speaker identity encoder $E_\mathrm{ID}$ and hierarchical pose decoder are trained with back-propagation from the above overall loss function.
\section{Experiments}
\begin{table*}[ht]
  \centering
  \begin{tabular}{lcccccccc}
    \toprule
     & \multicolumn{3}{c}{TED Gesture~\cite{yoon2020speech, yoon2019robots}} &  \multicolumn{3}{c}{TED-Expressive} \\
    \cmidrule(r){2-4} \cmidrule(r){5-7}
    Methods & FGD $\downarrow$ & BC $\uparrow$ & Diversity $\uparrow$ & FGD $\downarrow$ & BC $\uparrow$ & Diversity $\uparrow$ \\
    Ground Truth & 0 & 0.795 & 110.821 & 0 & 0.723 & 175.231\\
    \midrule
     Attention Seq2Seq~\cite{yoon2019robots}   & 18.154  & 0.186 & 92.176 & 54.920 & 0.155 & 122.693\\
     Speech2Gesture~\cite{ginosar2019learning}  & 19.254  & 0.764 & 98.095 & 54.650 & 0.714 & 142.489\\
     Joint Embedding~\cite{ahuja2019language2pose}   & 22.083  & 0.177 & 91.223 & 64.555 & 0.131 & 120.627\\
     Trimodal~\cite{yoon2020speech}   & 3.729 & 0.688 & 102.539 & 12.613 & 0.592 & 154.088\\
     \midrule 
     \textbf{HA2G (Ours)} & \textbf{3.072} & \textbf{0.769} & \textbf{108.086} & \textbf{5.306} & \textbf{0.715} & \textbf{173.899}\\
    \bottomrule
  \end{tabular}
  \caption{\textbf{The quantitative results on TED Gesture~\cite{yoon2020speech,yoon2019robots} and TED-Expressive.} We compare the proposed Hierarchical Audio-to-Gesture (\textbf{HA2G}) against recent SOTA methods~\cite{ahuja2019language2pose, ginosar2019learning, yoon2020speech, yoon2019robots} and ground truth under three metrics. For FGD the lower the better, and the higher the better for other metrics. Note that the FGD results of~\cite{ahuja2019language2pose, ginosar2019learning, yoon2020speech, yoon2019robots} on TED Gesture are reported from~\cite{yoon2020speech}.}
  \label{tbl:res}
  \vspace{-3mm}
\end{table*}
At the inference stage, we use speech audio as guidance while text is not needed. We further extract initial poses and speaker identity from a few reference images.
If the reference image is unavailable, we can sample initial poses from dataset and sample speaker identity from normal distribution to generate co-speech gestures since we constrain identity space with $\mathcal{L}_{\mathrm{KLD}}$. In this way, we can generate diverse gestures with multiple styles by sampling style vectors.

\subsection{Datasets and Annotation\textsuperscript{\ref{fotnt}}}
\label{dataset}
\noindent\textbf{TED Gesture.} TED Gesture dataset~\cite{yoon2020speech,yoon2019robots} is a large-scale English-language dataset for speech-driven motion synthesis, which contains 1,766 TED videos of different narrators covering various topics. The extracted 3D human skeletons, aligned English transcripts and speech audio are all available. Following~\cite{yoon2020speech}, we resample human poses with 15 FPS and sample the consecutive 34 frames with stride of 10 frames as input segments. We finally get 252,109 segments with length of 106.1h. In this dataset, human pose $\bm{p}$ is represented by direction vectors of 10 upper body joints.

\noindent\textbf{TED-Expressive.} The pose annotations of TED Gesture limit to 10 upper body keypoints without expressive co-speech finger movements. Hence, to harvest more detailed pose annotation as training data, we use the state-of-art 3D pose estimator ExPose~\cite{ExPose:2020} to extract 3D human skeleton as pseudo ground truth. In particular, we first annotate the 3D coordinates of 43 keypoints, including 13 upper body joints and 30 finger joints. Then we convert 3D coordinates into 42 unit direction vectors following~\cite{yoon2020speech} to represent each bone for eliminating the influence of various bone lengths in training data. In this way, our 3D representation is invariant to root joint motion and body shape. At the inference stage, the mean bone length over dataset is multiplied to the predicted bone vectors for visualized results. 

\subsection{Experimental Settings}
\label{exp-setting}
\noindent\textbf{Baselines.}
We compare our method with : (1) \textbf{Attention Seq2Seq}~\cite{yoon2019robots} which generates gestures from speech text by attention mechanism; (2) \textbf{Speech2Gesture}~\cite{ginosar2019learning} that takes the whole-length audio spectrogram as input and generates motion sequence with an encoder-decoder architecture and adversarial training scheme; (3) \textbf{Joint Embedding}~\cite{ahuja2019language2pose}, a representative method that maps the text and motion to the same embedding space and creates motion from description text; (4) \textbf{Trimodal}~\cite{yoon2020speech}, the state-of-art method that considers the trimodal context of text, audio and speaker identity to learn co-speech gestures. Note that some recent works~\cite{qian2021speech,li2021audio2gestures} lack open-source codes so far, thus we do not compare with them. All works are trained on the TED Gesture and TED-Expressive datasets for the same number of epochs with hyper-parameters optimized by grid search for best evaluation results. We also show the evaluation directly on the pseudo \textbf{Ground Truth} annotated in the dataset.

\noindent\textbf{Implementation Details.}\footnote{Please refer
to Supplementary Material for more details\label{fotnt}.} Following the settings of~\cite{yoon2020speech}, we set $N=34$ and $M=4$, so that the data are segmented into 34-frame sequences and the first 4 frames serve as reference frames. The number of joint $J$ is 10 for TED Gesture dataset and 43 for TED-Expressive dataset as mentioned in Sec.~\ref{dataset}. The audio encoder backbone is a ResNetSE34~\cite{chung2020defence} and the structure of text encoder $E_t$ is borrowed from~\cite{bai2018empirical}. The reference video frames are resized into $224 \times 224$, then passed into the speaker identity encoder $E_\mathrm{ID}$ with visual backbone of ResNet-18~\cite{he2016deep} to extract speaker identity. The raw audios are  converted to mel-spectrograms with FFT window size 1024, hop length 512. The word sequence is inserted with padding tokens to align with gestures. For each frame, 16 padded words and 0.25s mel-spectrogram with the target frame time-step in the middle are sampled as condition. The pose decoder is a cascaded 4-layer bi-directional GRU with a hidden size $d_s$ of 300 for each level of pose hierarchy. Empirically, we set $\tau=0.07$, $\epsilon=1000$, $d_a=32$, $\lambda_h=200$, $\lambda_p=0.1$, $\lambda_s=0.05$, $\lambda_k=0.1$, $\lambda_c=0.1$. The models are trained using Adam Optimizer with the learning rate of $1e-4$ on 1 GTX 1080Ti GPU.

\subsection{Quantitative Evaluation}
\noindent\textbf{Evaluation Metrics.}
We take the evaluation metrics that have been previously used in the co-speech gesture generation and music2dance for quantitative analysis.\\
\textbf{Fr\'echet Gesture Distance (FGD)} is used in~\cite{yoon2020speech} to measure how close the distribution of generated gesture is to the real one. Note that for the evaluation on TED Gesture dataset, we use the feature extractor provided in~\cite{yoon2020speech} for fair comparison. For the TED-Expressive dataset, we similarly train an auto-encoder on the TED-Expressive dataset and take the encoder part for feature extraction. FGD is calculated as the fr\'echet distance between the latent representations of real gesture and generated gesture.\\
\textbf{Beat Consistency Score (BC)} is a metric for motion-audio beat correlation as proposed in~\cite{li2021learn, li2021dancenet3d}. However, since the kinematic velocities vary from different joints, we propose to use the change of included angle between bones to track motion beats. Concretely, we calculate the mean absolute angle change (MAAC) of angle $\theta_j$ in adjacent frames by:
\begin{equation} 
    \label{eq:bc-mean}
    \mathrm{MAAC}(\theta_j) = \frac{\sum_{s=1}^S\sum_{t=1}^{T-1}\|\theta_{j, s, t+1} - \theta_{j, s, t}\|_1}{S * (T-1)},
\end{equation}
where $S$ is the total number of clips over dataset, $T$ is the number of frames for a clip and $\theta_{j, s, t}$ is included angle between the $j$-th and the ($j$+1)-th bone of the $s$-th clip at time-step $t$. In this way, the angle change rate of frame $t$ for the $s$-th clip is $\frac{1}{J-1}\sum_{j=1}^{J-1}(\|\theta_{j, s, t+1} - \theta_{j, s, t}\|_1/\mathrm{MAAC}(\theta_j))$. Then we extract the local optima whose first-order difference is higher than a threshold\textsuperscript{\ref{fotnt}} to get kinematic beats. We follow~\cite{li2021dancenet3d} to detect audio beat by onset strength~\cite{ellis2007beat} and compute the average distance between every audio beat and its nearest motion beat as Beat Consistency Score:
\begin{equation} 
    \label{eq:bc}
    \mathrm{BC} = \frac{1}{n}\sum_{i=1}^n\exp (-\frac{\min_{\forall  t_j^x\in B^x}\|t_i^x - t_j^y\|^2}{2\sigma^2}),
\end{equation}
where $B^x=\{t^x_i\}$ are the kinematic beats, $B^y=\{t^y_j\}$ are the audio beats and $\sigma$ is a parameter to normalize sequences that is empirically set to $0.1$ for experiments.

\noindent\textbf{Diversity} evaluates the variations among generated gestures corresponding to various inputs~\cite{NEURIPS2019_7ca57a9f}. Similarly, we use the same feature extractor in measuring FGD to map synthesized gestures into latent feature vectors and calculate the average feature distance for evaluation. Concretely, we randomly sample 60 speech audios from the test set to generate co-speech gestures and compute the average feature distance between 500 random combinated pairs.\\
\noindent\textbf{Evaluation Results.}
The results are shown in Table~\ref{tbl:res}. We can see that our \textbf{HA2G} framework outperforms existing methods on both datasets. Since our method establishes motion hierarchy and generates gestures in a coarse-to-fine manner, we can learn the diverse motion pattern of different human body parts and perform the best on FGD metric. Note that the improvement of FGD is smaller on TED Gesture dataset compared to TED-Expressive. This is due to the absence of finger information in TED Gesture dataset, which makes the motion hierarchy lower and the improvement brought by our hierarchical framework less significant. We can find that both Speech2Gesture~\cite{ginosar2019learning} and ours synthesize synchronous gestures to speech with high values on BC. But they tend to create unnatural poses and hence perform fair on FGD. In terms of Diversity, the discriminative feature extraction at multiple granularities enables us to excavate fine-grained audio-pose associations, thus capturing diverse speaking styles compared to baseline methods.

\begin{figure*}[t!]
    \centering
    \includegraphics[width=0.9\linewidth]{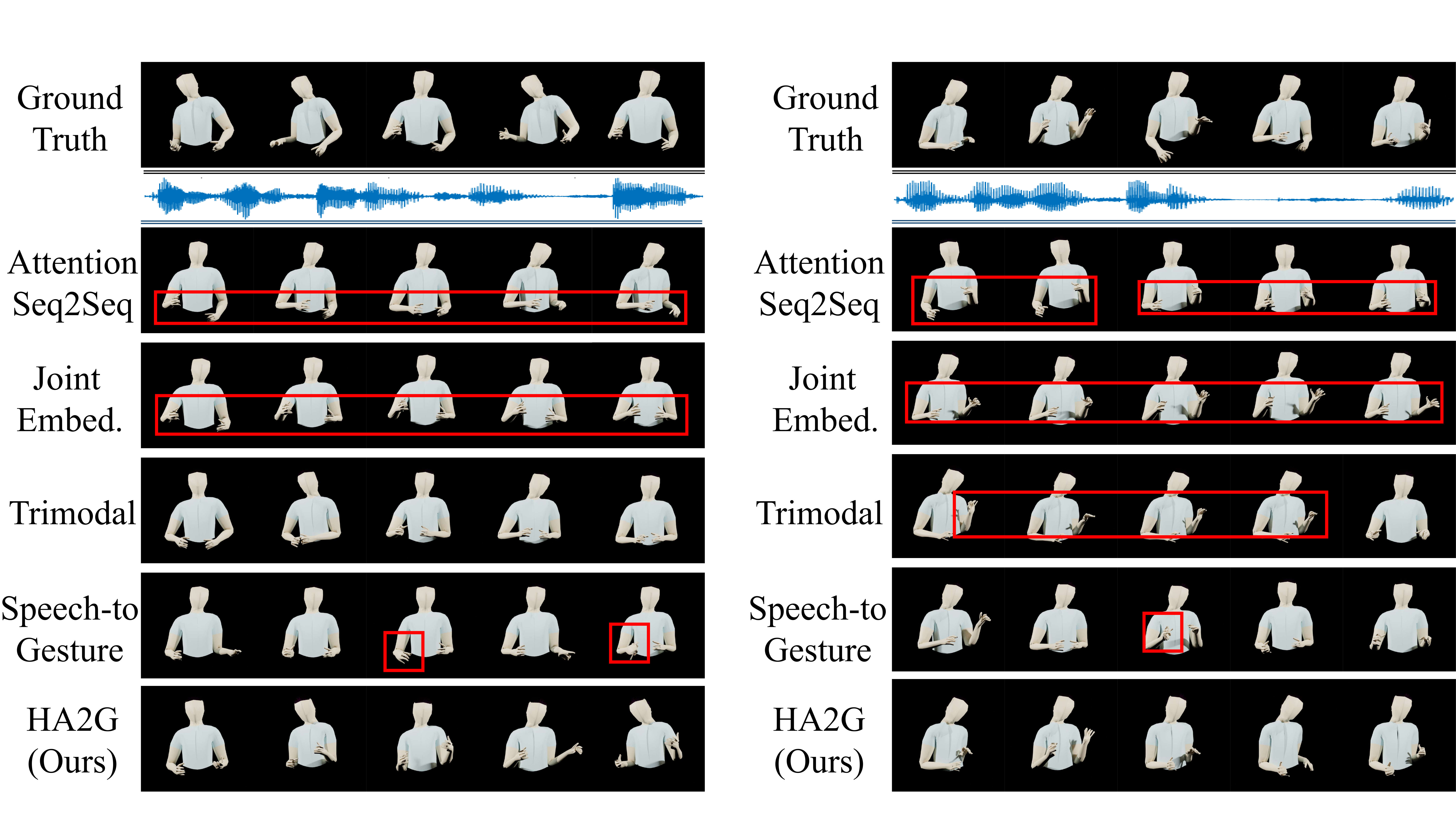}
    \vspace{-4mm}
    \caption{\textbf{The visualized results in two example clips.} We show the key frames of the generated motions from ground truth and baseline methods~\cite{ahuja2019language2pose, ginosar2019learning, yoon2020speech, yoon2019robots}. Please \textbf{zoom in for better visualization}. More high-resolution results can be found in the demo video.}
    \label{vis}
    \vspace{-2mm}
\end{figure*}

\begin{table*}
  \centering
  \begin{tabular}{ccccccc}
    \toprule
    Methods &
    GT &
    Seq2Seq~\cite{yoon2019robots}& 
    Joint.~\cite{ahuja2019language2pose}  & Tri.~\cite{yoon2020speech} & 
    S2G.~\cite{ginosar2019learning} & 
    \textbf{HA2G (Ours)} \\
    \midrule
    Naturalness & 4.16 & 1.36 & 1.52 & 3.66 & 2.88 &\textbf{4.13}\\
    Smoothness & 3.97 & \textbf{4.48} & 4.32 & 3.87 & 2.23 & 3.92\\
    Synchrony & 4.28 & 1.24 & 1.18 & 3.21 & 3.89 &\textbf{4.06}\\
        
    \bottomrule
  \end{tabular}
  \vspace{-3mm}
  \caption{\textbf{User study results on motion naturalness, smoothness and synchrony.} The rating is on a scale of 1-5, with the larger the better.}
  \label{table:userstudy}
  \vspace{-3mm}
\end{table*}

\subsection{Qualitative Evaluation}
\label{qualitative}
Subjective evaluation is crucial for judging the quality of results in generation tasks. Here we show the key frames comparison of our method against ground truth and SOTA baselines (as listed in Sec.~\ref{exp-setting}) in Fig.~\ref{vis}. For two cases, both Attention Seq2Seq~\cite{yoon2019robots} and Joint Embedding~\cite{ahuja2019language2pose} generate slow and invariant motions that are misaligned to speech as demonstrated in red rectangles of Fig.~\ref{vis}. While Trimodal~\cite{yoon2020speech} generates diverse gestures, the rigid motion pattern makes them mismatch to audio beats. For example, they stiffly move hands up and down with asynchronous beats to speech audio (see the red rectangle on the right). Both our method and Speech2Gesture~\cite{ginosar2019learning} create synchronous motions, but they synthesize unnatural poses, \textit{e.g.}, the twisted hands in both cases as highlighted in Fig.~\ref{vis}. The hierarchical cross-modal association against single-level design also leads to more diverse results than~\cite{ginosar2019learning}.

\noindent\textbf{User Study.}\footnote{Please refer to Supple. for more details about user study.} We conduct a user study on motion naturalness, smoothness and the generated co-speech gestures' synchrony to speech. In particular, we randomly sample 20 speech clips from test set of TED-Expressive to generate results for ground truth (tracked) annotations, baselines and our method. The study involves 24 participants. We adopt the widely-used Mean Opinion Scores (MOS) rating protocol, which requires the participants to rate three aspects of generated motions: (1) \textit{Naturalness}; (2) \textit{Smoothness}; (3) \textit{Synchrony between speech and generated gestures}. The rating is based on a scale of 1 to 5, with 5 being the most plausible and 1 being the least plausible.

The results are shown in Table~\ref{table:userstudy}. Since both Attention Seq2Seq~\cite{yoon2019robots} and JointEmbedding~\cite{ahuja2019language2pose} generate slow and near-stationary results, they score reasonably low on naturalness and synchrony, and trivially perform well on smoothness, which is even better than ground truth due to the motion jitter in ExPose annotation. Although Speech2Gesture~\cite{ginosar2019learning} performs well on synchrony, unnatural poses lead to fair results on naturalness and smoothness. Moreover, as our hierarchical design can capture fine-grained associations between multi-level features and diverse body parts, we score better than Trimodal~\cite{yoon2020speech} on all three aspects, with comparable results against ground truth. Note that to measure the disagreement on scoring among the participants, we also calculate the Fleiss's-Kappa\footnote {\href{https://en.wikipedia.org/wiki/Fleiss\%27\_kappa}{https://en.wikipedia.org/wiki/Fleiss\%27\_kappa}} statistic on 24 participants' ratings over all methods. The Fleiss-Kappa value is $0.837$, which is comparatively high and can be interpreted as ``almost perfect agreement''.
\subsection{Ablation Study}
In this section, we present ablation studies on two key modules proposed in our framework. We report the results implemented on the TED-Expressive dataset.

\noindent\textbf{Hierarchical Audio Learner.} To show the effect of multi-level audio feature in generating co-speech gesture, we conduct experiments on our model (1) $\bm{f}^\mathrm{low}_a$ only, which means we only use low-level feature from hierarchical audio encoder, \textit{i.e.}, the weight for low-level is set as 1 and weights for mid/high level features are set as 0 in Eq.~\ref{eq:speakerid}; (2) $\bm{f}^\mathrm{mid}_a$ only; (3) $\bm{f}^\mathrm{high}_a$ only; (4) w/o $\bm{f}^\mathrm{high}_{a-}$, which means we do not involve high level audio negative samples mentioned in Sec.~\ref{sec:3.2} for contrastive learning; (5) w/o $\bm{f}^\mathrm{low, mid}_{a-}$, which states the situation without cross-level negative samples; (6) w/o text, in this setting the input of speech text is not used, so we do not use the contrastive loss $\mathcal{L}_{\mathrm{multi}}$ for audio-text alignment and discriminative audio feature extraction. The results are shown in Table~\ref{tbl:ab1}, which indicates the efficacy of Hierarchical Audio Learner. Concretely, the only use of single-level audio feature fails to excavate information at multiple granularities, thus leading to degradation in performance. Besides, the contrastive learning strategy further improves performance since it achieves discriminative audio feature extraction with the self-supervision of audio-text alignment. More importantly, we find that our method \textbf{without} text outperforms Yoon \textit{et al.}~\cite{yoon2020speech} \textbf{with} the input of text. This demonstrates that the hierarchical design and coarse-to-fine generation manner can synthesize gestures of higher quality despite lack of text, enabling our method to handle general scenarios where video transcripts are unavailable.

\begin{table}
  \centering
  \begin{tabular}{l|ccc}
    \toprule
    Methods & FGD $\downarrow$ & BC $\uparrow$ & Diversity $\uparrow$ \\
    \midrule
     $\bm{f}^\mathrm{low}_a$ only & 6.588 & 0.704 & 171.482 \\
     $\bm{f}^\mathrm{mid}_a$ only & 7.212 & 0.682 & 168.223 \\
     $\bm{f}^\mathrm{high}_a$ only & 7.421 & 0.661 & 165.741 \\
     HA2G w/o $\bm{f}^\mathrm{high}_{a-}$& 7.982 & 0.652 & 163.649 \\
     HA2G w/o $\bm{f}^\mathrm{low, mid}_{a-}$ & 6.998 & 0.701 & 169.021 \\
     HA2G w/o text & 9.228 & 0.619 & 158.236 \\
     HA2G-ASR & 5.319 & \textbf{0.716} & 173.058\\
    %  HA2G(Ours) & 5.306 & 0.715 & 173.899\\
     \textbf{HA2G Full} & \textbf{5.306} & 0.715 & \textbf{173.899} \\
    \bottomrule
  \end{tabular}
  \caption{\textbf{Ablation study results of Hierarchical Audio Learner.}}
  \label{tbl:ab1}
   \vspace{-3mm}
\end{table}

\begin{table}
  \centering
  \begin{tabular}{l|ccc}
    \toprule
    Methods & FGD $\downarrow$ & BC $\uparrow$ & Diversity $\uparrow$ \\
    \midrule
     Holistic & 11.989 & 0.594 & 156.079 \\
     w/o hand hierarchy& 10.832 & 0.606 & 158.823 \\
     w/o body hierarchy& 5.882 & 0.709 & 173.066 \\
     Same audio $\bm{f}_{a}^h$ & 6.801 & 0.701 & 170.085 \\
     w/o $\mathcal{L}_{\mathrm{phy}}$ & 5.907 & 0.708 & 172.651 \\
     \textbf{HA2G Full} & \textbf{5.306} & \textbf{0.715} & \textbf{173.899} \\
    \bottomrule
  \end{tabular}
  \caption{\textbf{Ablation study results of Hierarchical Pose Inferer.}}
  \label{tbl:ab2}
   \vspace{-6mm}
\end{table}

Another ablation study relates to the Hierarchical Audio Learner is why we adopt contrastive learning strategy for discriminative feature extraction. We take inspiration from the fact that ASR models can semantically align text and audios, thus multi-level semantic information can be extracted from audio itself. However, the amount of data provided in the dataset is insufficient for training an expert ASR model, which leads to our choice of hierarchical contrastive design. For the ablation experiment, we use a well-trained ASR model~\cite{winata2020lightweight} as the audio encoder and generate co-speech gestures without contrastive strategy. The low, middle and high level features are also extracted from the backbone in a similar way as our method. We denote this variant of HA2G as HA2G-ASR. The comparisons on the TED-Expressive dataset are shown in the Table~\ref{tbl:ab1}. We can notice that the prior knowledge of pretrained ASR network prevents outlier predictions, which achieves competitive results compared to ours. This illustrates that using different levels of ASR features will benefit gesture generation. Note that the pretrained ASR network is trained on \textbf{a large amount of additional data}, while HA2G is trained with just a multi-level contrastive loss \textbf{without involving other pretrained networks and additional data}.

\noindent\textbf{Hierarchical Pose Inferer.} The experiments of Hierarchical Pose Inferer on our model contain: (1) Holistic, which means we do not use pose hierarchy and directly generate whole-body pose like previous methods~\cite{ginosar2019learning,yoon2020speech, ahuja2019language2pose, yoon2019robots}; (2) w/o hand hierarchy, where the hand poses are generated holistically while body hierarchy remains; (3) w/o body hierarchy, where body poses are generated holistically while hand hierarchy remains; (4) Same audio $\bm{f}_{a}^h$, which means we pass identical hierarchical audio features to each level of motion hierarchy, \textit{i.e.}, all columns of style coordinator $C$ are same in Eq.~\ref{eq:speakerid}; (5) w/o $\mathcal{L}_{\mathrm{phy}}$. Table~\ref{tbl:ab2} shows the results, which verify that Hierarchical Pose Inferer improves the performance. The pose hierarchy and distinct audio feature of each level enable the model to grasp fine-grained audio-pose associations of different body parts, making generated pose more vivid. The physical regularization $\mathcal{L}_{\mathrm{phy}}$ enhances FGD with more realistic human poses. Note that w/o body hierarchy outperforms w/o hand hierarchy. This is reasonable since the hand motion is more subtle, so hierarchical architecture's impact on hand is more significant.

\section{Discussion}
\label{conclusion}
\noindent\textbf{Conclusion.} In this paper, we propose a novel framework Hierarchical Audio-to-Gesture (\textbf{HA2G}) for co-speech gesture generation. We introduce Hierarchical Audio Learner with a contrastive learning strategy that extracts discriminative audio representations across semantic granularities. Then we propose Hierarchical Pose Inferer with a physical regularization to render the entire human pose gradually in a hierarchical manner. Extensive experiments demonstrate the superior performance of our proposed approach on co-speech gesture generation with high fidelity. 

\noindent\textbf{Limitation.} From the dataset perspective, our model is trained on an English-based corpus, which brings inductive bias on language. How to build a versatile model to generate co-speech gesture of diverse languages is a worthy direction for the community to explore.

\noindent\textbf{Acknowledgments}
This work has been supported by the Centre for Perceptual and Interactive Intelligence (CPII) Ltd under
the Innovation and Technology Fund, the RIE2020 Industry Alignment Fund–Industry Collaboration Projects (IAF-ICP) Funding Initiative, as well as cash and in-kind contribution from the industry partner(s).

%%%%%%%%% REFERENCES
{\small
\bibliographystyle{ieee_fullname}
\bibliography{egbib}

\begin{thebibliography}{10}\itemsep=-1pt

\bibitem{ahuja2020no}
Chaitanya Ahuja, Dong~Won Lee, Ryo Ishii, and Louis-Philippe Morency.
\newblock No gestures left behind: Learning relationships between spoken
  language and freeform gestures.
\newblock In {\em Proceedings of the 2020 Conference on Empirical Methods in
  Natural Language Processing: Findings}, pages 1884--1895, 2020.

\bibitem{ahuja2020style}
Chaitanya Ahuja, Dong~Won Lee, Yukiko~I Nakano, and Louis-Philippe Morency.
\newblock Style transfer for co-speech gesture animation: A multi-speaker
  conditional-mixture approach.
\newblock In {\em European Conference on Computer Vision}, pages 248--265.
  Springer, 2020.

\bibitem{ahuja2019language2pose}
Chaitanya Ahuja and Louis-Philippe Morency.
\newblock Language2pose: Natural language grounded pose forecasting.
\newblock In {\em 2019 International Conference on 3D Vision (3DV)}, pages
  719--728. IEEE, 2019.

\bibitem{aksan2019structured}
Emre Aksan, Manuel Kaufmann, and Otmar Hilliges.
\newblock Structured prediction helps 3d human motion modelling.
\newblock In {\em Proceedings of the IEEE/CVF International Conference on
  Computer Vision}, pages 7144--7153, 2019.

\bibitem{bai2018empirical}
Shaojie Bai, J~Zico Kolter, and Vladlen Koltun.
\newblock An empirical evaluation of generic convolutional and recurrent
  networks for sequence modeling.
\newblock {\em arXiv preprint arXiv:1803.01271}, 2018.

\bibitem{barsoum2018hp}
Emad Barsoum, John Kender, and Zicheng Liu.
\newblock Hp-gan: Probabilistic 3d human motion prediction via gan.
\newblock In {\em Proceedings of the IEEE conference on computer vision and
  pattern recognition workshops}, pages 1418--1427, 2018.

\bibitem{battan2021glocalnet}
Neeraj Battan, Yudhik Agrawal, Sai~Soorya Rao, Aman Goel, and Avinash Sharma.
\newblock Glocalnet: Class-aware long-term human motion synthesis.
\newblock In {\em Proceedings of the IEEE/CVF Winter Conference on Applications
  of Computer Vision}, pages 879--888, 2021.

\bibitem{bhattacharya2021text2gestures}
Uttaran Bhattacharya, Nicholas Rewkowski, Abhishek Banerjee, Pooja Guhan,
  Aniket Bera, and Dinesh Manocha.
\newblock Text2gestures: A transformer-based network for generating emotive
  body gestures for virtual agents.
\newblock In {\em 2021 IEEE Virtual Reality and 3D User Interfaces (VR)}, pages
  1--10. IEEE, 2021.

\bibitem{bowden2000learning}
Richard Bowden.
\newblock Learning statistical models of human motion.
\newblock In {\em IEEE Workshop on Human Modeling, Analysis and Synthesis,
  CVPR}, volume 2000. Citeseer, 2000.

\bibitem{brand2000style}
Matthew Brand and Aaron Hertzmann.
\newblock Style machines.
\newblock In {\em Proceedings of the 27th annual conference on Computer
  graphics and interactive techniques}, pages 183--192, 2000.

\bibitem{cao2019openpose}
Zhe Cao, Gines Hidalgo, Tomas Simon, Shih-En Wei, and Yaser Sheikh.
\newblock Openpose: realtime multi-person 2d pose estimation using part
  affinity fields.
\newblock {\em IEEE transactions on pattern analysis and machine intelligence},
  43(1):172--186, 2019.

\bibitem{cassell1999speech}
Justine Cassell, David McNeill, and Karl-Erik McCullough.
\newblock Speech-gesture mismatches: Evidence for one underlying representation
  of linguistic and nonlinguistic information.
\newblock {\em Pragmatics \& cognition}, 7(1):1--34, 1999.

\bibitem{cassell1994animated}
Justine Cassell, Catherine Pelachaud, Norman Badler, Mark Steedman, Brett
  Achorn, Tripp Becket, Brett Douville, Scott Prevost, and Matthew Stone.
\newblock Animated conversation: rule-based generation of facial expression,
  gesture \& spoken intonation for multiple conversational agents.
\newblock In {\em Proceedings of the 21st annual conference on Computer
  graphics and interactive techniques}, pages 413--420, 1994.

\bibitem{cassell2004beat}
Justine Cassell, Hannes~H{\"o}gni Vilhj{\'a}lmsson, and Timothy Bickmore.
\newblock Beat: the behavior expression animation toolkit.
\newblock In {\em Life-Like Characters}, pages 163--185. Springer, 2004.

\bibitem{chen2020comprises}
Lele Chen, Guofeng Cui, Ziyi Kou, Haitian Zheng, and Chenliang Xu.
\newblock What comprises a good talking-head video generation?: A survey and
  benchmark.
\newblock {\em arXiv preprint arXiv:2005.03201}, 2020.

\bibitem{chen2019hierarchical}
Lele Chen, Ross~K Maddox, Zhiyao Duan, and Chenliang Xu.
\newblock Hierarchical cross-modal talking face generation with dynamic
  pixel-wise loss.
\newblock In {\em Proceedings of the IEEE/CVF Conference on Computer Vision and
  Pattern Recognition}, pages 7832--7841, 2019.

\bibitem{ExPose:2020}
Vasileios Choutas, Georgios Pavlakos, Timo Bolkart, Dimitrios Tzionas, and
  Michael~J. Black.
\newblock Monocular expressive body regression through body-driven attention.
\newblock In {\em European Conference on Computer Vision (ECCV)}, 2020.

\bibitem{chung2020defence}
Joon~Son Chung, Jaesung Huh, Seongkyu Mun, Minjae Lee, Hee~Soo Heo, Soyeon
  Choe, Chiheon Ham, Sunghwan Jung, Bong-Jin Lee, and Icksang Han.
\newblock In defence of metric learning for speaker recognition.
\newblock {\em arXiv preprint arXiv:2003.11982}, 2020.

\bibitem{ellis2007beat}
Daniel Ellis.
\newblock Beat tracking by dynamic programming.
\newblock {\em Journal of New Music Research}, 36:51--60, 03 2007.

\bibitem{ferstl2020adversarial}
Ylva Ferstl, Michael Neff, and Rachel McDonnell.
\newblock Adversarial gesture generation with realistic gesture phasing.
\newblock {\em Computers \& Graphics}, 89:117--130, 2020.

\bibitem{galata2001learning}
Aphrodite Galata, Neil Johnson, and David Hogg.
\newblock Learning variable-length markov models of behavior.
\newblock {\em Computer Vision and Image Understanding}, 81(3):398--413, 2001.

\bibitem{gao20192}
Ruohan Gao and Kristen Grauman.
\newblock 2.5 d visual sound.
\newblock In {\em Proceedings of the IEEE Conference on Computer Vision and
  Pattern Recognition}, pages 324--333, 2019.

\bibitem{ghosh2017learning}
Partha Ghosh, Jie Song, Emre Aksan, and Otmar Hilliges.
\newblock Learning human motion models for long-term predictions.
\newblock In {\em 2017 International Conference on 3D Vision (3DV)}, pages
  458--466. IEEE, 2017.

\bibitem{ginosar2019learning}
Shiry Ginosar, Amir Bar, Gefen Kohavi, Caroline Chan, Andrew Owens, and
  Jitendra Malik.
\newblock Learning individual styles of conversational gesture.
\newblock In {\em Proceedings of the IEEE/CVF Conference on Computer Vision and
  Pattern Recognition}, pages 3497--3506, 2019.

\bibitem{habibie2021learning}
Ikhsanul Habibie, Weipeng Xu, Dushyant Mehta, Lingjie Liu, Hans-Peter Seidel,
  Gerard Pons-Moll, Mohamed Elgharib, and Christian Theobalt.
\newblock Learning speech-driven 3d conversational gestures from video.
\newblock {\em arXiv preprint arXiv:2102.06837}, 2021.

\bibitem{hasegawa2018evaluation}
Dai Hasegawa, Naoshi Kaneko, Shinichi Shirakawa, Hiroshi Sakuta, and Kazuhiko
  Sumi.
\newblock Evaluation of speech-to-gesture generation using bi-directional lstm
  network.
\newblock In {\em Proceedings of the 18th International Conference on
  Intelligent Virtual Agents}, pages 79--86, 2018.

\bibitem{he2016deep}
Kaiming He, Xiangyu Zhang, Shaoqing Ren, and Jian Sun.
\newblock Deep residual learning for image recognition.
\newblock In {\em 2016 IEEE Conference on Computer Vision and Pattern
  Recognition (CVPR)}, pages 770--778, 2016.

\bibitem{hernandez2019human}
Alejandro Hernandez, Jurgen Gall, and Francesc Moreno-Noguer.
\newblock Human motion prediction via spatio-temporal inpainting.
\newblock In {\em Proceedings of the IEEE/CVF International Conference on
  Computer Vision}, pages 7134--7143, 2019.

\bibitem{2017GANs}
M. Heusel, H. Ramsauer, T. Unterthiner, B. Nessler, and S. Hochreiter.
\newblock Gans trained by a two time-scale update rule converge to a local nash
  equilibrium.
\newblock In {\em Neural Information Processing Systems (NIPS)}, 2017.

\bibitem{hinton2012deep}
Geoffrey Hinton, Li Deng, Dong Yu, George~E Dahl, Abdel-rahman Mohamed, Navdeep
  Jaitly, Andrew Senior, Vincent Vanhoucke, Patrick Nguyen, Tara~N Sainath,
  et~al.
\newblock Deep neural networks for acoustic modeling in speech recognition: The
  shared views of four research groups.
\newblock {\em IEEE Signal processing magazine}, 29(6):82--97, 2012.

\bibitem{holden2016deep}
Daniel Holden, Jun Saito, and Taku Komura.
\newblock A deep learning framework for character motion synthesis and editing.
\newblock {\em ACM Transactions on Graphics (TOG)}, 2016.

\bibitem{huang2012robot}
Chien-Ming Huang and Bilge Mutlu.
\newblock Robot behavior toolkit: generating effective social behaviors for
  robots.
\newblock In {\em 2012 7th ACM/IEEE International Conference on Human-Robot
  Interaction (HRI)}, pages 25--32. IEEE, 2012.

\bibitem{huang2020dance}
Ruozi Huang, Huang Hu, Wei Wu, Kei Sawada, Mi Zhang, and Daxin Jiang.
\newblock Dance revolution: Long-term dance generation with music via
  curriculum learning.
\newblock {\em arXiv preprint arXiv:2006.06119}, 2020.

\bibitem{huber1992robust}
Peter~J Huber.
\newblock Robust estimation of a location parameter.
\newblock In {\em Breakthroughs in statistics}, pages 492--518. Springer, 1992.

\bibitem{ishi2018}
Carlos~T. Ishi, Daichi Machiyashiki, Ryusuke Mikata, and Hiroshi Ishiguro.
\newblock A speech-driven hand gesture generation method and evaluation in
  android robots.
\newblock {\em IEEE Robotics and Automation Letters}, 3(4):3757--3764, 2018.

\bibitem{ji2021audio-driven}
Xinya Ji, Hang Zhou, Kaisiyuan Wang, Wayne Wu, Chen~Change Loy, Xun Cao, and
  Feng Xu.
\newblock Audio-driven emotional video portraits.
\newblock In {\em Proceedings of the IEEE Conference on Computer Vision and
  Pattern Recognition (CVPR)}, 2021.

\bibitem{kucherenko2020gesticulator}
Taras Kucherenko, Patrik Jonell, Sanne van Waveren, Gustav~Eje Henter, Simon
  Alexandersson, Iolanda Leite, and Hedvig Kjellstr\"{o}m.
\newblock Gesticulator: A framework for semantically-aware speech-driven
  gesture generation.
\newblock In {\em Proceedings of the 2020 International Conference on
  Multimodal Interaction}, ICMI '20, page 242–250, New York, NY, USA, 2020.
  Association for Computing Machinery.

\bibitem{lee2019talking}
Gilwoo Lee, Zhiwei Deng, Shugao Ma, Takaaki Shiratori, Siddhartha~S Srinivasa,
  and Yaser Sheikh.
\newblock Talking with hands 16.2 m: A large-scale dataset of synchronized
  body-finger motion and audio for conversational motion analysis and
  synthesis.
\newblock In {\em Proceedings of the IEEE/CVF International Conference on
  Computer Vision}, pages 763--772, 2019.

\bibitem{NEURIPS2019_7ca57a9f}
Hsin-Ying Lee, Xiaodong Yang, Ming-Yu Liu, Ting-Chun Wang, Yu-Ding Lu,
  Ming-Hsuan Yang, and Jan Kautz.
\newblock Dancing to music.
\newblock In H. Wallach, H. Larochelle, A. Beygelzimer, F. d\textquotesingle
  Alch\'{e}-Buc, E. Fox, and R. Garnett, editors, {\em Advances in Neural
  Information Processing Systems}, volume~32. Curran Associates, Inc., 2019.

\bibitem{levine2010gesture}
Sergey Levine, Philipp Kr{\"a}henb{\"u}hl, Sebastian Thrun, and Vladlen Koltun.
\newblock Gesture controllers.
\newblock In {\em ACM SIGGRAPH 2010 papers}, pages 1--11. 2010.

\bibitem{li2021dancenet3d}
Buyu Li, Yongchi Zhao, and Lu Sheng.
\newblock Dancenet3d: Music based dance generation with parametric motion
  transformer.
\newblock {\em arXiv preprint arXiv:2103.10206}, 2021.

\bibitem{0Convolutional}
C. Li, Z. Zhang, W.~S. Lee, and G.~H. Lee.
\newblock Convolutional sequence to sequence model for human dynamics.
\newblock In {\em IEEE/CVF Conference on Computer Vision and Pattern
  Recognition}, 2018.

\bibitem{li2021audio2gestures}
Jing Li, Di Kang, Wenjie Pei, Xuefei Zhe, Ying Zhang, Zhenyu He, and Linchao
  Bao.
\newblock Audio2gestures: Generating diverse gestures from speech audio with
  conditional variational autoencoders.
\newblock {\em arXiv preprint arXiv:2108.06720}, 2021.

\bibitem{li2021learn}
Ruilong Li, Shan Yang, David~A Ross, and Angjoo Kanazawa.
\newblock Learn to dance with aist++: Music conditioned 3d dance generation.
\newblock {\em arXiv preprint arXiv:2101.08779}, 2021.

\bibitem{lin2017feature}
Tsung-Yi Lin, Piotr Doll{\'a}r, Ross Girshick, Kaiming He, Bharath Hariharan,
  and Serge Belongie.
\newblock Feature pyramid networks for object detection.
\newblock In {\em Proceedings of the IEEE conference on computer vision and
  pattern recognition}, pages 2117--2125, 2017.

\bibitem{liu2022visual}
Xian Liu, Rui Qian, Hang Zhou, Di Hu, Weiyao Lin, Ziwei Liu, Bolei Zhou, and
  Xiaowei Zhou.
\newblock Visual sound localization in the wild by cross-modal interference
  erasing.
\newblock {\em arXiv preprint arXiv:2202.06406}, 2022.

\bibitem{liu2022semantic}
Xian Liu, Yinghao Xu, Qianyi Wu, Hang Zhou, Wayne Wu, and Bolei Zhou.
\newblock Semantic-aware implicit neural audio-driven video portrait
  generation.
\newblock {\em arXiv preprint arXiv:2201.07786}, 2022.

\bibitem{loehr2012temporal}
Daniel~P Loehr.
\newblock Temporal, structural, and pragmatic synchrony between intonation and
  gesture.
\newblock {\em Laboratory phonology}, 2012.

\bibitem{marsella2013virtual}
Stacy Marsella, Yuyu Xu, Margaux Lhommet, Andrew Feng, Stefan Scherer, and Ari
  Shapiro.
\newblock Virtual character performance from speech.
\newblock In {\em Proceedings of the 12th ACM SIGGRAPH/Eurographics Symposium
  on Computer Animation}, pages 25--35, 2013.

\bibitem{mcneill2011hand}
David McNeill.
\newblock {\em Hand and mind}.
\newblock De Gruyter Mouton, 2011.

\bibitem{nefian2002dynamic}
Ara~V Nefian, Luhong Liang, Xiaobo Pi, Xiaoxing Liu, and Kevin Murphy.
\newblock Dynamic bayesian networks for audio-visual speech recognition.
\newblock {\em EURASIP Journal on Advances in Signal Processing},
  2002(11):1--15, 2002.

\bibitem{newell2016stacked}
Alejandro Newell, Kaiyu Yang, and Jia Deng.
\newblock Stacked hourglass networks for human pose estimation.
\newblock In {\em European conference on computer vision}, pages 483--499.
  Springer, 2016.

\bibitem{prajwal2020lip}
KR Prajwal, Rudrabha Mukhopadhyay, Vinay~P Namboodiri, and CV Jawahar.
\newblock A lip sync expert is all you need for speech to lip generation in the
  wild.
\newblock In {\em Proceedings of the 28th ACM International Conference on
  Multimedia}, pages 484--492, 2020.

\bibitem{pullen2000animating}
Katherine Pullen and Christoph Bregler.
\newblock Animating by multi-level sampling.
\newblock In {\em Proceedings Computer Animation 2000}, pages 36--42. IEEE,
  2000.

\bibitem{qian2021speech}
Shenhan Qian, Zhi Tu, YiHao Zhi, Wen Liu, and Shenghua Gao.
\newblock Speech drives templates: Co-speech gesture synthesis with learned
  templates.
\newblock {\em arXiv preprint arXiv:2108.08020}, 2021.

\bibitem{gentle}
Ochshorn Robert and Hawkin Max.
\newblock Gentle: A forced aligner.
\newblock 2016.

\bibitem{ronneberger2015u}
Olaf Ronneberger, Philipp Fischer, and Thomas Brox.
\newblock U-net: Convolutional networks for biomedical image segmentation.
\newblock In {\em International Conference on Medical image computing and
  computer-assisted intervention}, pages 234--241. Springer, 2015.

\bibitem{sadoughi2015msp}
Najmeh Sadoughi, Yang Liu, and Carlos Busso.
\newblock Msp-avatar corpus: Motion capture recordings to study the role of
  discourse functions in the design of intelligent virtual agents.
\newblock In {\em 2015 11th IEEE International Conference and Workshops on
  Automatic Face and Gesture Recognition (FG)}, volume~7, pages 1--6. IEEE,
  2015.

\bibitem{salem2012generation}
Maha Salem, Stefan Kopp, Ipke Wachsmuth, Katharina Rohlfing, and Frank Joublin.
\newblock Generation and evaluation of communicative robot gesture.
\newblock {\em International Journal of Social Robotics}, 4(2):201--217, 2012.

\bibitem{salem2011friendly}
Maha Salem, Katharina Rohlfing, Stefan Kopp, and Frank Joublin.
\newblock A friendly gesture: Investigating the effect of multimodal robot
  behavior in human-robot interaction.
\newblock In {\em 2011 Ro-Man}, pages 247--252. IEEE, 2011.

\bibitem{Shi_2016_CVPR}
Wenzhe Shi, Jose Caballero, Ferenc Huszar, Johannes Totz, Andrew~P. Aitken, Rob
  Bishop, Daniel Rueckert, and Zehan Wang.
\newblock Real-time single image and video super-resolution using an efficient
  sub-pixel convolutional neural network.
\newblock In {\em Proceedings of the IEEE Conference on Computer Vision and
  Pattern Recognition (CVPR)}, June 2016.

\bibitem{tian2021cyclic}
Yapeng Tian, Di Hu, and Chenliang Xu.
\newblock Cyclic co-learning of sounding object visual grounding and sound
  separation.
\newblock In {\em Proceedings of the IEEE/CVF Conference on Computer Vision and
  Pattern Recognition}, 2021.

\bibitem{tian2020unified}
Yapeng Tian, Dingzeyu Li, and Chenliang Xu.
\newblock Unified multisensory perception: Weakly-supervised audio-visual video
  parsing.
\newblock In {\em European Conference on Computer Vision}. Springer, 2020.

\bibitem{tian2018audio}
Yapeng Tian, Jing Shi, Bochen Li, Zhiyao Duan, and Chenliang Xu.
\newblock Audio-visual event localization in unconstrained videos.
\newblock In {\em Proceedings of the European Conference on Computer Vision
  (ECCV)}, pages 247--263, 2018.

\bibitem{tolins2016multimodal}
Jackson Tolins, Kris Liu, Yingying Wang, Jean E~Fox Tree, Marilyn Walker, and
  Michael Neff.
\newblock A multimodal motion-captured corpus of matched and mismatched
  extravert-introvert conversational pairs.
\newblock In {\em Proceedings of the Tenth International Conference on Language
  Resources and Evaluation (LREC'16)}, pages 3469--3476, 2016.

\bibitem{van1998persona}
Susanne Van~Mulken, Elisabeth Andre, and Jochen M{\"u}ller.
\newblock The persona effect: how substantial is it?
\newblock In {\em People and computers XIII}, pages 53--66. Springer, 1998.

\bibitem{villegas2018neural}
Ruben Villegas, Jimei Yang, Duygu Ceylan, and Honglak Lee.
\newblock Neural kinematic networks for unsupervised motion retargetting.
\newblock In {\em Proceedings of the IEEE Conference on Computer Vision and
  Pattern Recognition}, pages 8639--8648, 2018.

\bibitem{2014Gesture}
P. Wagner, Z. Malisz, and S. Kopp.
\newblock Gesture and speech in interaction: An overview.
\newblock {\em Speech Communication}, 57:209--232, 2014.

\bibitem{wang2020deep}
Jingdong Wang, Ke Sun, Tianheng Cheng, Borui Jiang, Chaorui Deng, Yang Zhao,
  Dong Liu, Yadong Mu, Mingkui Tan, Xinggang Wang, et~al.
\newblock Deep high-resolution representation learning for visual recognition.
\newblock {\em IEEE transactions on pattern analysis and machine intelligence},
  2020.

\bibitem{wei2019motion}
Mao Wei, Liu Miaomiao, Salzemann Mathieu, and Li Hongdong.
\newblock Learning trajectory dependencies for human motion prediction.
\newblock In {\em ICCV}, 2019.

\bibitem{winata2020lightweight}
Genta~Indra Winata, Samuel Cahyawijaya, Zhaojiang Lin, Zihan Liu, and Pascale
  Fung.
\newblock Lightweight and efficient end-to-end speech recognition using
  low-rank transformer.
\newblock In {\em ICASSP 2020-2020 IEEE International Conference on Acoustics,
  Speech and Signal Processing (ICASSP)}, pages 6144--6148. IEEE, 2020.

\bibitem{xu2021visually}
Xudong Xu, Hang Zhou, Ziwei Liu, Bo Dai, Xiaogang Wang, and Dahua Lin.
\newblock Visually informed binaural audio generation without binaural audios.
\newblock In {\em Proceedings of the IEEE conference on computer vision and
  pattern recognition (CVPR)}, 2021.

\bibitem{yan2019convolutional}
Sijie Yan, Zhizhong Li, Yuanjun Xiong, Huahan Yan, and Dahua Lin.
\newblock Convolutional sequence generation for skeleton-based action
  synthesis.
\newblock In {\em Proceedings of the IEEE/CVF International Conference on
  Computer Vision}, pages 4394--4402, 2019.

\bibitem{yoon2020speech}
Youngwoo Yoon, Bok Cha, Joo-Haeng Lee, Minsu Jang, Jaeyeon Lee, Jaehong Kim,
  and Geehyuk Lee.
\newblock Speech gesture generation from the trimodal context of text, audio,
  and speaker identity.
\newblock {\em ACM Transactions on Graphics (TOG)}, 39(6):1--16, 2020.

\bibitem{yoon2019robots}
Youngwoo Yoon, Woo-Ri Ko, Minsu Jang, Jaeyeon Lee, Jaehong Kim, and Geehyuk
  Lee.
\newblock Robots learn social skills: End-to-end learning of co-speech gesture
  generation for humanoid robots.
\newblock In {\em 2019 International Conference on Robotics and Automation
  (ICRA)}, pages 4303--4309. IEEE, 2019.

\bibitem{yu2020audio}
Jianwei Yu, Shi-Xiong Zhang, Jian Wu, Shahram Ghorbani, Bo Wu, Shiyin Kang,
  Shansong Liu, Xunying Liu, Helen Meng, and Dong Yu.
\newblock Audio-visual recognition of overlapped speech for the lrs2 dataset.
\newblock In {\em ICASSP 2020-2020 IEEE International Conference on Acoustics,
  Speech and Signal Processing (ICASSP)}, pages 6984--6988. IEEE, 2020.

\bibitem{zhao2019sound}
Hang Zhao, Chuang Gan, Wei-Chiu Ma, and Antonio Torralba.
\newblock The sound of motions.
\newblock In {\em Proceedings of the IEEE International Conference on Computer
  Vision}, pages 1735--1744, 2019.

\bibitem{zhou2019talking}
Hang Zhou, Yu Liu, Ziwei Liu, Ping Luo, and Xiaogang Wang.
\newblock Talking face generation by adversarially disentangled audio-visual
  representation.
\newblock In {\em AAAI Conference on Artificial Intelligence (AAAI)}, 2019.

\bibitem{zhou2021pose}
Hang Zhou, Yasheng Sun, Wayne Wu, Chen~Change Loy, Xiaogang Wang, and Ziwei
  Liu.
\newblock Pose-controllable talking face generation by implicitly modularized
  audio-visual representation.
\newblock In {\em Proceedings of the IEEE/CVF Conference on Computer Vision and
  Pattern Recognition}, 2021.

\bibitem{zhou2020sep}
Hang Zhou, Xudong Xu, Dahua Lin, Xiaogang Wang, and Ziwei Liu.
\newblock Sep-stereo: Visually guided stereophonic audio generation by
  associating source separation.
\newblock In {\em Proceedings of the European Conference on Computer Vision
  (ECCV)}, 2020.

\bibitem{zhou2020makelttalk}
Yang Zhou, Xintong Han, Eli Shechtman, Jose Echevarria, Evangelos Kalogerakis,
  and Dingzeyu Li.
\newblock Makelttalk: speaker-aware talking-head animation.
\newblock {\em ACM Transactions on Graphics (TOG)}, 39(6):1--15, 2020.

\bibitem{zhou2020generative}
Yi Zhou, Jingwan Lu, Connelly Barnes, Jimei Yang, Sitao Xiang, et~al.
\newblock Generative tweening: Long-term inbetweening of 3d human motions.
\newblock {\em arXiv preprint arXiv:2005.08891}, 2020.

\end{thebibliography}
}

\clearpage
\appendix
\twocolumn[{%
\begin{minipage}{\textwidth}
   \null
   \vspace*{0.375in}
   \begin{center}
      {\Large \bf Supplemental Document: Learning Hierarchical Cross-Modal Association \\for Co-Speech Gesture Generation \par}
   \end{center}
   \vspace*{0.375in}
\end{minipage}
\vspace{-6mm}}]

\renewcommand{\thesection}{\Alph{section}}

%%%%%%%%% ABSTRACT
\section{More Details about Dataset}
\noindent\textbf{Choice of Data and Data Collection.} Many learning-based approaches use motion and gesture training data captured in a MoCap studio with complex motion capture systems~\cite{lee2019talking,levine2010gesture}. They can acquire more accurate human motion data compared to automatic annotations on internet videos. However, such methods have the following drawbacks: 1) Owing to the high cost of MoCap data, it is hard to build a large-scale corpus of data covering various speaking contents and styles. For example, the length of MSP-AVATAR~\cite{sadoughi2015msp} and Personality Dyads Corpus~\cite{tolins2016multimodal} are less than 3h. 2) When capturing co-speech gesture data in the studio, the actors/actresses are asked to deliberately talk with their arms and hands moving, which contributes to the unnaturalness and exaggeration of captured motion data. Therefore, we follow the previous works~\cite{ginosar2019learning,yoon2020speech,yoon2019robots} to collect internet videos and annotate 3D human pose as pseudo ground truth for later training. Specifically, Ginosar \textit{et al.}~\cite{ginosar2019learning} and Habibie \textit{et al.}~\cite{habibie2021learning} use a speaker-specific gesture dataset of a very small number of speakers, \textit{i.e.}, 10 speakers in~\cite{ginosar2019learning} and 6 speakers in~\cite{habibie2021learning}, making them unable to transfer to general speaking styles. TED Gesture dataset is proposed by Yoon \textit{et al.}~\cite{yoon2020speech} which contains over 1,700 TED talks covering diverse topics and speaker styles. Following Yoon \textit{et al.}~\cite{yoon2020speech}, we propose to build our TED-Expressive dataset based on the raw videos of TED talks. Differently, since the flexible finger movement matters a lot when people talk, we add the information of finger keypoints for more expressive co-speech gesture dataset establishment. We collect internet videos from the official TED channel on YouTube.\footnote{We obey the TED Talks Team's Creative Commons License (CC BY-NC-ND 4.0 International) by referencing all the video links shown in our papers. We sincerely thank the permission of the TED Talks Team for using the videos, audios and transcripts in this paper.} We finally get 1,764 videos and their corresponding text transcripts.

\noindent\textbf{Pose Annotation and Post-Processing.} To get the reliable pseudo ground truth of co-speech human upper body pose with finger keypoints, we leverage the state-of-art 3D human motion estimator ExPose~\cite{ExPose:2020} for annotation. Similar to the step of~\cite{yoon2019robots}, we segment videos into smaller shots by their scenes and annotate the 2D human pose of each frame by OpenPose~\cite{cao2019openpose}. With the 2D pose prior provided by OpenPose, we use ExPose~\cite{ExPose:2020} to annotate 3D upper body keypoints. Concretely, we use 43 keypoints, including 13 upper body joints (spine, neck, nose, left/right eyes, ears, shoulders, elbows and wrists, totally $13 = 3 + 5 * 2$) and 30 finger joints (3 joints for each finger, totally $30 = 3 * 5 * 2$). Then we select shots of interest under the following conditions: 1) the above mentioned 43 keypoints of speaker are visible for more than 50\% frames of a clip; 2) the speaker should not remain almost still in the whole shot, \textit{i.e.}, the variance of motion is quite small; 3) the clip is longer than 5s. The statistics of TED Gesture and TED-Expressive dataset are recorded in Table~\ref{statistic}. For the TED Gesture dataset, we randomly split the segments into the 80\% training set, 10\% validation set, 10\% test set and finally get 199,384; 26,795; and 25,930 segments in each partition. 
\begin{table*}[ht]
    \centering
  \begin{tabular}{lcccc}
    \toprule
    Statistics & \# of Videos & Interest Shots Length & \# of Segments & Interest Ratio \\
    \midrule
     TED Gesture~\cite{yoon2020speech,yoon2019robots} & 1,766  & 106.1h & 252,109 & 25\%\\
     TED-Expressive & 1,764 & 100.8h & 240,447 & 21\%\\
    \bottomrule
  \end{tabular}
  \caption{Statistics of the TED Gesture and TED-Expressive dataset.}
    \label{statistic}
\end{table*}

\begin{figure}
    \centering
    \includegraphics[width=1\linewidth]{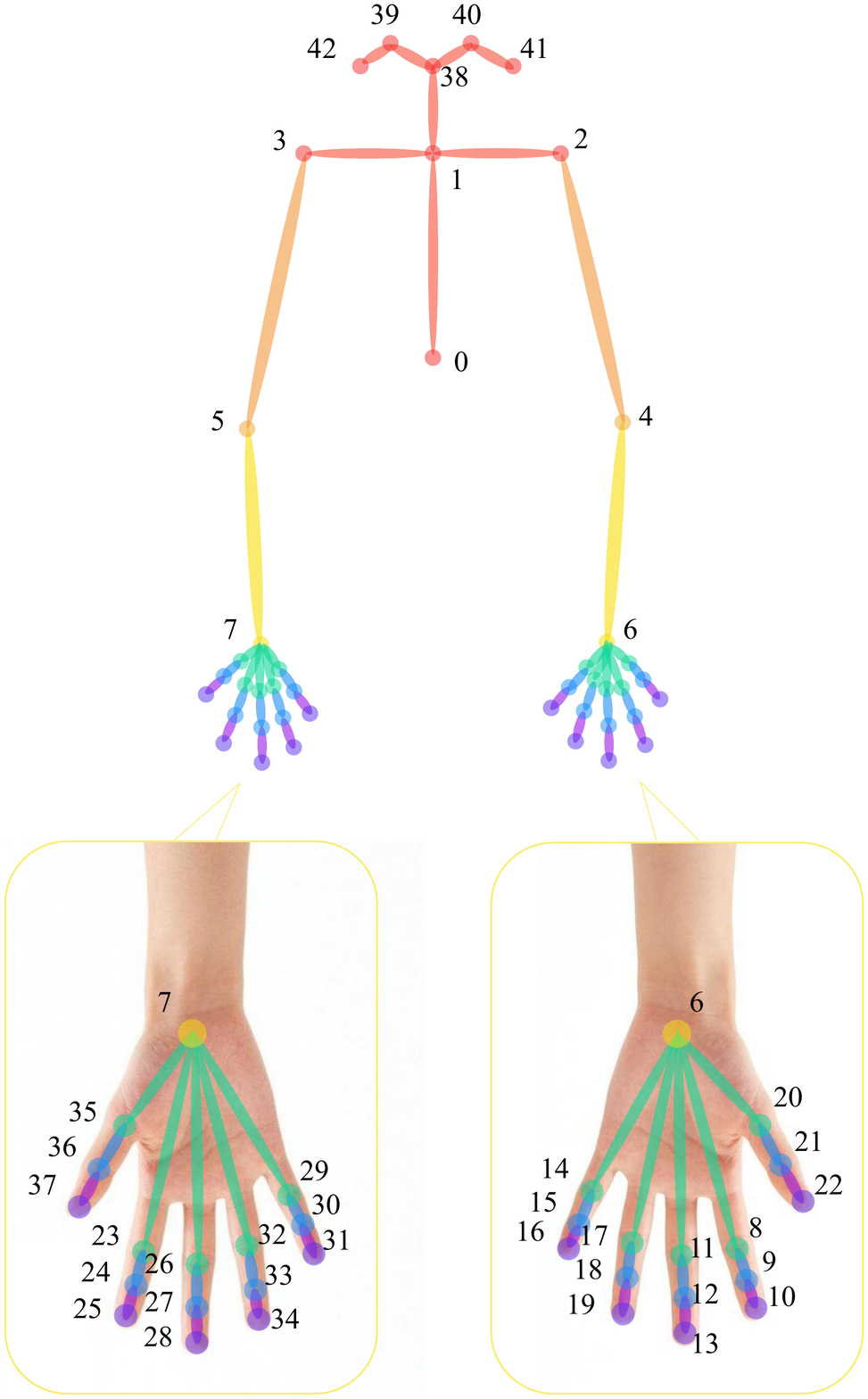}
    \vspace{-6mm}
    \caption{\textbf{The detailed 3D keypoints output annotation by ExPose~\cite{ExPose:2020}.} In particular, we annotate 43 upper body keypoints, including: spine (0), neck (1), left shoulder (2), right shoulder (3), left elbow (4), right elbow (5), left wrist (6), right wrist (7), left index (8, 9, 10), left middle (11, 12, 13), left pinky (14, 15, 16), left ring (17, 18, 19), left thumb (20, 21, 22), right index (23, 24, 25), right middle (26, 27, 28), right pinky (29, 30, 31), right ring (32, 33, 34), right thumb (35, 36, 37), nose (38), right eye (39), left eye (40), right ear (41), left ear (42). Note that the holistic upper body with keypoints index is shown at the top of figure, the zoom-in images of left hand and right hand with detailed annotations are shown at the bottom.}
    \label{annotation}
\end{figure}

\noindent\textbf{Pose Representation and Quality.} After the filtering process, we effectively eliminate the influence of bone length by normalizing them into 42 unit directional vectors to represent each bone. Such 3D representation is invariant to root joint motion and body shape, thus making it more stable in the training phase. At the inference stage, the mean bone length over the training set is multiplied to predicted bone vectors for visualized results. The whole pipeline is automated, which facilitates us to build a large corpus of co-speech gesture dataset. Figure~\ref{annotation} shows the correspondence between keypoint index and joints. We can see that there are totally 43 annotated upper body keypoints, which are then transformed into 42 unit direction vectors as mentioned above.

As the pose annotations serve as pseudo ground truth in our pipeline, the quality of annotations is crucial for training. However, since the pose representation is 3D, we can not follow Ginosar \textit{et al.}~\cite{ginosar2019learning} to evaluate the quality of annotations by automatic pipeline against human annotations. But the high performance of ExPose on benchmark datasets and our filtering algorithm guarantee that the data quality is good enough for utilization. Please refer to ExPose~\cite{ExPose:2020} for the detailed quantitative 3D pose estimation results on benchmark dataset. Overall, we use the open-source code of Trimodal~\cite{yoon2020speech}, OpenPose~\cite{cao2019openpose} and ExPose~\cite{ExPose:2020} following their licenses\footnote{ExPose License:
\href{https://github.com/vchoutas/expose/blob/master/LICENSE}{https://github.com/vchoutas/expose/blob/master/LICE\\NSE};
OpenPose License: \href{https://github.com/CMU-Perceptual-Computing-Lab/openpose/blob/master/LICENSE}{https://github.com/CMU-Perceptual-Computing\\-Lab/openpose/blob/master/LICENSE};
Trimodal: \href{https://github.com/ai4r/Gesture-Generation-from-Trimodal-Context/blob/master/LICENSE.md}{https://github.com/ai4r\\/Gesture-Generation-from-Trimodal-Context/blob/master/LICENSE.md}}.

\noindent\textbf{Speech Audio Pre-Processing.} The speech audios accompanied TED videos are raw waveforms, which are processed to 16kHZ and convert to mel-spectrograms as 2D time-frequency representations for more compact information preservation. The FFT window size is 1024 and the hop length is 512.

\noindent\textbf{Speech Text Pre-Processing.} We collect speech text input with the transcripts of TED videos. Then, the onset timestamps of each word are extracted by the Gentle forced aligner~\cite{gentle} to insert padding tokens. For example, for the speech text ``Good morning everyone'', if there is a short pause between the word ``morning'' and ``everyone'', then the padded word sequence is ``Good morning $\diamond$ $\diamond$ everyone'' as padded by Gentle if the time of this sentence is 5. Following the process of~\cite{yoon2020speech}, the padded word sequences are transformed into word vectors of 300 dimensions through a word embedding layer.

\section{Architecture Details}
\vspace{-1mm}
\noindent\textbf{Audio Encoder $E_a$.} The audio encoder is a ResNetSE34 borrowed from~\cite{chung2020defence}. Specifically, we define the features output from ResNet Stage-2 as shallow feature map, features output from ResNet Stage-3 as middle feature map, features output from ResNet Stage-4 as deep feature map. Then, a series of upsampling, convolution, batchnorm and linear layers transform corresponding audio feature maps into the same size. When the input audio mel-spectrogram of size $1 \times 128 \times 70$, the channel dimension and frequency, time resolutions of different level features $\bm{f}^{\mathrm{low}}_a$, $\bm{f}^{\mathrm{mid}}_a$ and $\bm{f}^{\mathrm{high}}_a\in \mathbb{R}^{32}$ with their corresponding operations are shown in Table~\ref{audio}. The detailed feature dimensions after each operation are shown in Table~\ref{audio-specific}. In this way, the hierarchical audio features are transformed into the same shape and the time dimension is exactly the frame number of a clip, \textit{i.e.}, 34 in our experiment, which is convenient for RNN-based model to take information of each time step as input. After the linear blending of multi-level features, hierarchical audio features for different levels of body parts are established and finally feed to cascaded bi-GRU for pose generation in a coarse-to-fine manner.

\begin{table*}[t]
    \centering
    \begin{tabular}{c|c|c|c|c}
    \hline
        Feature Map & Output & Shape & Operation & Feature\\
        \hline
        Input & - & $1 \times 128 \times 70$ & Pre-Conv & - \\
        Shallow & Stage-2 & $64 \times 64 \times 35$ & Conv2d, ReLU, BN, FC & $\bm{f}^{\mathrm{low}}_a$ \\
        Middle & Stage-3 & $128 \times 32 \times 18$ & PixelShuffle, Conv2d, ReLU, BN, FC & $\bm{f}^{\mathrm{mid}}_a$ \\
        Deep & Stage-4 & $256 \times 16 \times 9$  & PixelShuffle, Conv2d, ReLU, BN, FC & $\bm{f}^{\mathrm{high}}_a$ \\
        \hline
    \end{tabular}
    \caption{\textbf{Definitions of multi-level audio feature maps and transformed features on ResNetSE34.} The shallow/middle/deep feature maps are from the output of ResNetSE34's stage2/3/4. Then, they are transformed by a series of operations to low/mid/high-level audio features, respectively.}
    \label{audio}
    \vspace{-3mm}
\end{table*}

\begin{table}[t]
  \centering{
  \begin{tabular}{|c|c|c|c|}
    \hline
           \multicolumn{4}{|c|}{Shallow Festure Map from Stage-2} \\
    \hline
           \multicolumn{2}{|c|}{Operations} & \multicolumn{2}{|c|}{Feature Map Shapes} \\
    
    \hline
           \multicolumn{2}{|c|}{Input} & \multicolumn{2}{|c|}{$64 \times 64 \times 35$} \\
    %\hline
          \multicolumn{2}{|c|}{Conv2d (64, 2, 1)} & \multicolumn{2}{|c|}{$64 \times 63 \times 34$} \\
    %\hline
           \multicolumn{2}{|c|}{ReLU, BatchNorm2d (64)} & \multicolumn{2}{|c|}{$64 \times 63 \times 34$} \\
    %\hline
            \multicolumn{2}{|c|}{Reshape} & \multicolumn{2}{|c|}{$4032 \times 34$} \\
    
          \multicolumn{2}{|c|}{FC (4032, 32)} & \multicolumn{2}{|c|}{$32 \times 34$} \\
    \hline
           \multicolumn{4}{|c|}{Middle Festure Map from Stage-3} \\
    \hline
           \multicolumn{2}{|c|}{Operations} & \multicolumn{2}{|c|}{Feature Map Shapes} \\
    
    \hline
           \multicolumn{2}{|c|}{Input} & \multicolumn{2}{|c|}{$128 \times 32 \times 18$} \\
           \multicolumn{2}{|c|}{PixelShuffle (2)} & \multicolumn{2}{|c|}{$32 \times 64 \times 36$} \\
    %\hline
          \multicolumn{2}{|c|}{Conv2d (32, 3, 1)} & \multicolumn{2}{|c|}{$32 \times 62 \times 34$} \\
    %\hline
           \multicolumn{2}{|c|}{ReLU, BatchNorm2d (32)} & \multicolumn{2}{|c|}{$32 \times 62 \times 34$} \\
    %\hline
    \multicolumn{2}{|c|}{Reshape} & \multicolumn{2}{|c|}{$1984 \times 34$} \\
          \multicolumn{2}{|c|}{FC (1984, 32)} & \multicolumn{2}{|c|}{$32 \times 34$} \\
     \hline
           \multicolumn{4}{|c|}{Deep Festure Map from Stage-4} \\
    \hline
           \multicolumn{2}{|c|}{Operations} & \multicolumn{2}{|c|}{Feature Map Shapes} \\
    
    \hline
           \multicolumn{2}{|c|}{Input} & \multicolumn{2}{|c|}{$256 \times 16 \times 9$} \\
           \multicolumn{2}{|c|}{PixelShuffle (4)} & \multicolumn{2}{|c|}{$16 \times 64 \times 36$} \\
    %\hline
          \multicolumn{2}{|c|}{Conv2d (16, 3, 1)} & \multicolumn{2}{|c|}{$16 \times 62 \times 34$} \\
    %\hline
           \multicolumn{2}{|c|}{ReLU, BatchNorm2d (16)} & \multicolumn{2}{|c|}{$16 \times 62 \times 34$} \\
           \multicolumn{2}{|c|}{Reshape} & \multicolumn{2}{|c|}{$992 \times 34$} \\
    %\hline
          \multicolumn{2}{|c|}{FC (992, 32)} & \multicolumn{2}{|c|}{$32 \times 34$} \\
    %\hline
    \hline
  \end{tabular}
  }
  \caption{\textbf{Detailed feature shape after specific operations for the multi-level audio feature extraction.} ${}^\dagger$Note that in the table, Conv2d ($c$, $k$, $s$) means the output of the convolution is $c$, kernel size is $k$ and the stride is $s$; ReLU, BatchNorm2d ($c$) means the relu and batch-norm operation on the feature of channel size $c$; Reshape operation combines the channel and frequency dimension together; FC ($i$, $o$) means the fully connected linear layer whose input dimension is $i$ and output dimension is $o$; PixelShuffle ($r$) means the pixel shuffle operation~\cite{Shi_2016_CVPR} with resolution $r$. Specifically, PixelShuffle ($r$) transforms a feature map of shape $(r^2C) \times H \times W$ into $C \times (rH) \times (rW)$.}
  \label{audio-specific}
\end{table}

\noindent\textbf{Text Encoder $E_t$.} With the speech text pre-processing mentioned above, the word sequences are transformed into word vectors. Next, these word vectors are encoded by an off-the-shelf temporal convolutional text encoder~\cite{bai2018empirical}. The text encoder $E_t$ is 4-layered, the receptive field is 16 padded words centered at the current time step and the output dimension of text feature $\bm{f}_t = E_t(\mathbf{t})$ is 32. In this way, the high-level audio feature and text feature at time-step $t$ are both of dimension 32, which enables the later contrastive learning strategy to leverage the natural audio-text correspondence for achieving discriminative cross-modal feature extraction.

\noindent\textbf{Speaker Identity Encoder $E_\mathrm{ID}$.} The speaker identity encoder network $E_\mathrm{ID}$ uses the standard ResNet-18-S5 model. And we do not load ImageNet pretrained weights. We also remove \emph{Global Average Pooling} (GAP) and the final classifier to only leave the visual backbone for visual feature extraction. The input of the encoder is the first $M$ reference frame images of a clip $\{I_1, \dots, I_M\}$ and the output is speaker identity embedding $\bm{f}_{id}=E_\mathrm{ID}(I_1, \dots, I_M) \in \mathbb{R}^{18}$. Then through a linear layer of FC (18, 18) and softmax function, $\bm{f}_{id}$ is transformed into the style coordinator $C\in \mathbb{R}^{3 \times H}$, where $\sum^3_{i=1} C[i, h] = 1$. In this way, the speaker identity can affect hierarchical audio feature weight on motion hierarchies.

\noindent\textbf{Motion Hierarchy Establishment.} The skeleton of human body is like a tree structure, where the father-joint carries the child-joint to move. To effectively learn the dynamic patterns of different human body parts, we propose to detach the joints from human body ends (fingers) to the main structure (spine) step-by-step and build a motion hierarchy of totally 6 hierarchies: 1) Nose, neck, spine and left/right eye, ear, shoulder; 2) Add left and right elbow; 3) Add left and right wrist; 4) Add left and right finger's first joint; 5) Add left and right finger's second joint; 6) Add left and right finger's third joint. Note that we do not separate more detailed motion hierarchy inside the skeleton of human head. Because there is hardly internal movement in the skeleton of head part, \textit{i.e.}, nose, left/right eye and ear. The dynamic among the keypoints of head part is quite trivial under our setting, so it is unreasonable to separate them into different levels of motion hierarchies. 

\noindent\textbf{Coarse-to-Fine Pose Generator.} According to the established motion hierarchy, the number of keypoints for 6 hierarchies is 9, 11, 13, 23, 33, 43. Since the poses are processed into 3D unit directional vectors representation, the pose dimension of 6 hierarchies is 24, 30, 36, 66, 96, 126. The cascaded hierarchical pose generator contains 6 bi-directional GRU, with the input of corresponding hierarchy pose and audio feature and the hidden size of 300. Note that for the first motion hierarchy, the poses of the first $M$ frames serve as initial poses and are denoted as $\hat{\mathbf{p}}^0=\{\bm{p}^0_1, ..., \bm{p}^0_M, 0, ..., 0\}$; for the later hierarchies, the output from the last pose hierarchy is leveraged to initialize corresponding keypoints. Therefore, $d_s=300$, $d_a=32$, $d_p^0 = 24$, $d_p^1 = 24$, $d_p^2 = 30$, $d_p^3 = 36$, $d_p^4 = 66$, $d_p^5 = 96$ and $d_p^6 = 126$ for the parameters $W^h$ and $\bm{b}^h$ of hierarchical GRU in the Eqn. 4 of main document. In this way, the next pose hierarchy can generate pose with information from the last level of pose, facilitating fine-grained correspondences between audio sequence and co-speech gestures in a coarse-to-fine manner. The last layer's output $\hat{\mathbf{p}}^H$ from the hierarchy is our desired result.

\noindent\textbf{GAN Discriminator $D$.}
The architecture of discriminator $D$ is borrowed from~\cite{yoon2020speech}, with its detailed network design in the Table~\ref{discriminator}.

\noindent\textbf{Pose Auto-Encoder.} Since the generation is a multi-modality problem, it is difficult to use evaluation metrics like L1 distance or L2 distance to judge whether the generated result is good or not. Fr\'echet Inception Distance (FID)~\cite{2017GANs} is widely leveraged to evaluate the image generation quality. It firstly pre-train a feature extractor to extract image latent features, then calculates the Fr\'echet distance between the distributions of the latent feature space of real and generated images. The feature vectors contain more information about characteristics, which is more perceptually plausible than raw pixel space. Based on this, Yoon \textit{et al.}~\cite{yoon2020speech} propose a similar evaluation metric Fr\'echet Gesture Distance (FGD) to evaluate gesture quality. 

To further evaluate the pose with expressive finger movements, we train a pose auto-encoder with 43 keypoints on TED-Expressive dataset. The auto-encoder firstly maps 34-frame poses into latent dimension of 128, and then reconstruct them. The detailed structure is borrowed from~\cite{yoon2020speech} and recorded in Table~\ref{ae}.

\section{Training Stage and Inference Stage}
At the training stage, the speech audios, transcripts and reference frames are all needed. The speech audio $\mathbf{a}$ is encoded by the hierarchical audio encoder $E_a$ to get multi-level audio features $\bm{f}^{\mathrm{low}}_a$, $\bm{f}^{\mathrm{mid}}_a$ and $\bm{f}^{\mathrm{high}}_a$. The speech transcript $\mathbf{t}$ is encoded by $E_t$ into text features $\bm{f}_{t}$, which are then used by contrastive learning strategy to achieve more discriminative audio feature extraction. Therefore, speech text takes an auxiliary effect in our proposed framework. The reference frames $\mathbf{I}=(I_1, \dots, I_M)$ are encoded by $E_{\mathrm{ID}}$ to represent speaker's identity $\bm{f}_{id}$, which is then transformed to style coordinator $C$ for feature blending. Besides, reference frames are also used to extract initial poses and finally feed into cascaded bi-GRU to generate co-speech gestures in a coarse-to-fine manner.

At the inference stage, the speech transcript is not needed. \textbf{This is the reason why we do not involve the variable $\mathbf{t}$ in the Eq. 1} in our main text. If the reference frames and initial poses are available, we can follow the whole pipeline to generate gestures. For the situations where reference frames and initial poses are unavailable, we can sample a style vector from normal distribution to serve as speaker identity $\bm{f}_{id}$. Then we can sample an arbitrary sequence of initial poses from the dataset to generate the gestures. 

\begin{table*}[t]
  \centering{
  \begin{tabular}{|c|c|c|c|c|c|}
    \hline
           \multicolumn{6}{|c|}{Discriminator $D$} \\
    \hline
           \multicolumn{2}{|c|}{Feature} & \multicolumn{2}{|c|}{Feature Shapes}& \multicolumn{2}{|c|}{Operations} \\
    
    \hline
           \multicolumn{2}{|c|}{Input} & \multicolumn{2}{|c|}{$34 \times 126$}& \multicolumn{2}{|c|}{Transpose (0, 1)} \\
    \hline
           \multicolumn{2}{|c|}{Pre-Conv Layer-1} & \multicolumn{2}{|c|}{$126 \times 34$}& \multicolumn{2}{|c|}{Conv1d (126, 16, 3), BatchNorm1d (16), LeakyReLU (0.2)} \\
    \hline
           \multicolumn{2}{|c|}{Pre-Conv Layer-2} & \multicolumn{2}{|c|}{$16 \times 32$}& \multicolumn{2}{|c|}{Conv1d (16, 8, 3), BatchNorm1d (8), LeakyReLU (0.2)} \\
    \hline
           \multicolumn{2}{|c|}{Pre-Conv Layer-3} & \multicolumn{2}{|c|}{$8 \times 30$}& \multicolumn{2}{|c|}{Conv1d (8, 8, 3), Transpose (0, 1)} \\
    \hline
           \multicolumn{2}{|c|}{Bi-Directional GRU} & \multicolumn{2}{|c|}{$28 \times 8$}& \multicolumn{2}{|c|}{Bi-Directional GRU (8, 64)} \\
    \hline
           \multicolumn{2}{|c|}{FC-1} & 
           \multicolumn{2}{|c|}{$28 \times 64$}& \multicolumn{2}{|c|}{FC (64, 1), Squeeze(1)} \\
    \hline
           \multicolumn{2}{|c|}{FC-2} & 
           \multicolumn{2}{|c|}{$28$}& \multicolumn{2}{|c|}{FC (28, 1), Sigmoid} \\
    \hline
           \multicolumn{2}{|c|}{Output} & 
           \multicolumn{2}{|c|}{$1$}& \multicolumn{2}{|c|}{-} \\
    \hline
  \end{tabular}
  }
  \caption{\textbf{Detailed structure and feature shape of Discriminator $D$.} ${}^\dagger$Note that in the table, the meanings of contents in operations column are: Conv1d (in\_channels, out\_channels, kernel\_size), BatchNorm1d (feature\_dim), LeakyReLU (alpha), Transpose (axis1, axis2), Bi-directional GRU (in\_size, hidden\_size), FC (in\_size, out\_size), Squeeze (axis), Sigmoid.}
  \label{discriminator}
\end{table*}

\begin{table*}[t]
  \centering{
  \begin{tabular}{|c|c|c|c|c|c|}
    \hline
           \multicolumn{6}{|c|}{Pose Encoder} \\
    \hline
           \multicolumn{2}{|c|}{Feature} & \multicolumn{2}{|c|}{Feature Shapes}& \multicolumn{2}{|c|}{Operations} \\
    
    \hline
           \multicolumn{2}{|c|}{Input} & \multicolumn{2}{|c|}{$34 \times 126$}& \multicolumn{2}{|c|}{Transpose (0, 1)} \\
    \hline
           \multicolumn{2}{|c|}{Layer-1} & \multicolumn{2}{|c|}{$126 \times 34$}& \multicolumn{2}{|c|}{Conv1d (32, 3, 1), BatchNorm1d (32), LeakyReLU (0.2)} \\
    \hline
           \multicolumn{2}{|c|}{Layer-2} & \multicolumn{2}{|c|}{$32 \times 32$}& \multicolumn{2}{|c|}{Conv1d (64, 3, 1), BatchNorm1d (64), LeakyReLU (0.2)} \\
    \hline
           \multicolumn{2}{|c|}{Layer-3} & \multicolumn{2}{|c|}{$64 \times 30$}& \multicolumn{2}{|c|}{Conv1d (64, 4, 2), BatchNorm1d (64), LeakyReLU (0.2)} \\
    \hline
           \multicolumn{2}{|c|}{Layer-4} & \multicolumn{2}{|c|}{$64 \times 14$}& \multicolumn{2}{|c|}{Conv1d (32, 3, 1)} \\
    \hline
           \multicolumn{2}{|c|}{Out1} & 
           \multicolumn{2}{|c|}{$32 \times 12$}& \multicolumn{2}{|c|}{Flatten, FC (384, 256), BatchNorm1d (256), LeakyReLU (0.2)} \\
    \hline
           \multicolumn{2}{|c|}{Out2} & 
           \multicolumn{2}{|c|}{$256$}& \multicolumn{2}{|c|}{FC (256, 128), BatchNorm1d (128), LeakyReLU (0.2), FC (128, 128)} \\
    \hline
           \multicolumn{2}{|c|}{Latent} & 
           \multicolumn{2}{|c|}{$128$}& \multicolumn{2}{|c|}{-} \\
    \hline
    \multicolumn{6}{|c|}{Pose Decoder} \\
    \hline
           \multicolumn{2}{|c|}{Feature} & \multicolumn{2}{|c|}{Feature Shapes}& \multicolumn{2}{|c|}{Operations} \\
    
    \hline
           \multicolumn{2}{|c|}{Input} & \multicolumn{2}{|c|}{$128$}& \multicolumn{2}{|c|}{FC (128, 64), BatchNorm1d (64), LeakyReLU (0.2), FC (64, 136)} \\
    \hline
           \multicolumn{2}{|c|}{reshape} & \multicolumn{2}{|c|}{$136$}& \multicolumn{2}{|c|}{Reshape (4, 34)} \\
    \hline
           \multicolumn{2}{|c|}{Layer-1} & \multicolumn{2}{|c|}{$4 \times 34$}& \multicolumn{2}{|c|}{ConvTranspose1d (32, 3, 1), BatchNorm1d (32), LeakyReLU (0.2)} \\
    \hline
           \multicolumn{2}{|c|}{Layer-2} & \multicolumn{2}{|c|}{$32 \times 36$}& \multicolumn{2}{|c|}{ConvTranspose1d (32, 3, 1), BatchNorm1d (32), LeakyReLU (0.2)} \\
    \hline
           \multicolumn{2}{|c|}{Layer-3} & \multicolumn{2}{|c|}{$32 \times 38$}& \multicolumn{2}{|c|}{Conv1d (32, 3, 1)} \\
    \hline
           \multicolumn{2}{|c|}{Layer-4} & \multicolumn{2}{|c|}{$32 \times 36$}& \multicolumn{2}{|c|}{Conv1d (126, 3, 1), Transpose(0, 1)} \\
    \hline
           \multicolumn{2}{|c|}{Pose} & 
           \multicolumn{2}{|c|}{$34 \times 126$}& \multicolumn{2}{|c|}{-} \\
    \hline
  \end{tabular}
  }
  \caption{\textbf{Detailed structure and feature shape of Pose Auto-Encoder.} ${}^\dagger$Note that in the table, Conv1d/ConvTranspose1d ($c$, $k$, $s$) means the output of the convolution/transpose-convolution is $c$, kernel size is $k$ and the stride is $s$; LeakyReLU, BatchNorm1d ($c$) means the leaky-relu and batch-norm operation on the feature of channel size $c$.}
  \label{ae}
\end{table*}

\section{Statistics in Physical Constraint}
Previous methods on co-speech gesture generation mostly fail to consider human physics constraints, which contributes to unnatural pose and incoherent results. Therefore, we propose to add restrictions on the included angle between bones to ensure reasonable human pose. Concretely, the pose is represented as bone direction vector, which is rendered as $\bm{p} = [\bm{d}_1, \bm{d}_2, \cdots , \bm{d}_{J-1}]$ and $J$ is the total number of joints. For the $j$-th bone vector $\bm{d}_j \in \mathbb{R}^{3}$ and the ($j$+1)-th bone vector $\bm{d}_{j+1} \in \mathbb{R}^{3}$, we can compute their included angle $\theta_j$ by the arc-cosine function on their cosine value. Since there is no benchmark dataset with accurate finger keypoints annotations \emph{under co-speech settings}, we 
use the hand pose estimator ExPose~\cite{ExPose:2020} to annotate the TED-Expressive dataset. With the pseudo ground truth, we can calculate the mean and variance of each angle, which later serve as the mean and variance of Gaussian distribution. The loss function for the physics constraint is the log-likelihood function:
\begin{equation} 
    \mathcal{L}_{\mathrm{phy}} = -\log\prod_{j=1}^{J-1}\mathcal{N}(\theta_j; \mu_j, \sigma_j^2)=-\sum_{j=1}^{J-1}\log\mathcal{N}(\theta_j; \mu_j, \sigma_j^2),
\end{equation}
where $\theta_j = \arccos{\frac{\bm{d}_j \cdot \bm{d}_{j+1}}{\|\bm{d}_j\| \|\bm{d}_{j+1}\|}}$ is the $j$-th angle value, $\mu_j$ and $\sigma_j^2$ are the mean and variance of the $j$-th angle respectively. We illustrate in Table~\ref{stat} the means and variances of the included angles (0-180 degrees) around two important joints. In particular, we use $\theta_s$ to denote the included angle around the shoulder joint and $\theta_e$ for the included angle around the elbow joint. Although some angles may not strictly follow the Gaussian distribution, the intention of physical constraint is to prevent outlier predictions. Thus the assumption of Gaussian distribution could play the role in regularizing generated poses. The ablation study in Table 4 (the setting of "w/o $\mathcal{L}_{\mathrm{phy}}$") shows the effectiveness of such a constraint.

\begin{table}[H]
    \centering
  \begin{tabular}{lcccc}
    \toprule
    Statistics & Left $\theta_s$ & Left $\theta_e$ & Right $\theta_s$ & Right $\theta_e$\\
    \midrule
     mean($^{\circ}$) & 116.6 & 75.1 & 127.1 & 85.3\\
     var($^{\circ}$) & 9.01 & 7.30 & 7.53 & 7.22\\
    \bottomrule
  \end{tabular}
  \caption{Statistics of important important joint angles.}
    \label{stat}
\end{table}

\section{Analysis on Beat Consistency Score Metric}
Beat Consistency Score (BC) is a metric adapted by us for motion-audio beat correlation. Previous methods detect motion beats by finding the local optima of kinematic velocity~\cite{li2021dancenet3d}, while we propose to utilize the change of included angle between bones to track motion beats. The main reasons are two-fold: 1) previous methods are under the setting of music2dance, where human body involves a global body translation in a large scale. In other word, all of the human's body joints move fast when people dance and the velocity quickly drops when they stop to match a music beat. While in our co-speech gesture settings, the arms are comparatively still and the fingers are more flexible, so their moving scales vary a lot, we can not directly sum up them. 2) Compared to using the shifts of keypoints, we propose to use the included angle to detect motion beat. This is because the human body follow a tree structure. If the arm moves, hand and wrist will follow the movement of arm, which is similar to the process of orbital revolution and self rotation: the father-joint carry the child-joint to move like the orbital revolution and the internal movement of child-joint resembles self rotation. Therefore, directly calculating the Euclidean distance for each joint is unreasonable.

\begin{figure}[t]
    \centering
    \includegraphics[width=0.8\linewidth]{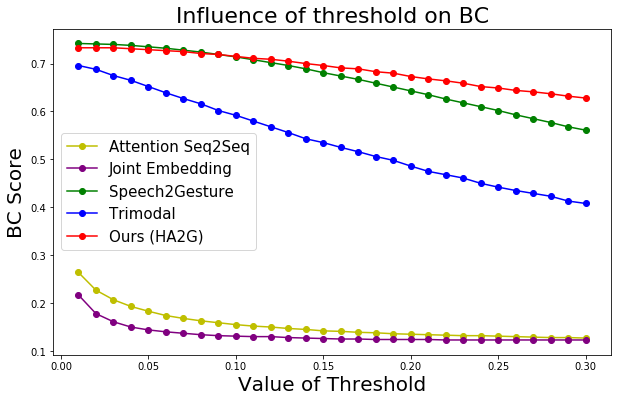}
    \caption{\textbf{The influence of threshold on Beat Consistency (BC) Score metric.} We present the BC value of baseline methods and ours under the threshold of range 0.01 to 0.3 with step size of 0.01.}
    \label{bc}
\end{figure}

After calculating mean absolute angle change (MAAC) of angle $\theta_j$, we can calculate the sum angle change rate of a certain frame $t$ for the $n$-th clip as:
\begin{equation} 
\frac{1}{J-1}\sum_{j=1}^{J-1}\frac{\|\theta_{j, n, t+1} - \theta_{j, n, t}\|_1}{\mathrm{MAAC}(\theta_j)}.
\end{equation}
Then we propose to extract the kinematic beat through filtering the angle change rate by following conditions: 1) The angle change rate is a local optimum, \textit{e.g.}, the angle change rate of 9, 10, 11 time-step is 0.2, 0.1, 0.2, respectively. Then the time-step 10 is a local optimum. 2) The difference of the local optima with either side time-step is larger than a threshold. This is to filter the trivial situation where angle change rates are almost the same during a period of time and guarantee a sudden change of angle change rate as motion beat. For example, the angle change rate of 8, 9, 10, 11, 12 time-step is 0.11, 0.1, 0.11, 0.1, 0.11. It improper to take the time-step 9 and 11 as motion beat. The threshold controls what extent of angle change rate difference is perceived it as a motion beat. A very low threshold will detect the near-stationary motion sequence as many motion beats if there are many trivial beats of type 2 mentioned in the last paragraph. A very high threshold will ignore the normal motion beat. We present the influence of threshold over all baseline method in Fig.~\ref{bc}. We can see that our method can achieve superior performance on BC metric with high robustness to threshold compared to baseline methods. Note that both Attention Seq2Seq~\cite{yoon2019robots} and Joint Embedding~\cite{ahuja2019language2pose} show low value of BC Score over all threshold, which also proves that they fail to generate results that are synchronous to speech since their gestures are almost still. Although Speech2Gesture~\cite{ginosar2019learning} shows higher performance on low threshold, they match the trivial beats and perform lower than our method on normal thresholds.

\section{Choice of Speaker Identity Extraction}
We leverage RGB frames rather than poses for identity information extraction. The appearances of different identities would vary significantly. Though dynamic information can hardly be inferred from the $M$ inputs, our method focuses more on the appearance information like the speaker's height, age and nationality. Inferring speaking styles from appearances or identities only has also been proven effective in Yoon \emph{et al.}~\cite{yoon2020speech}. The only speaker-related information we can access is the initial frames, thus we are trying to make the best use of them.

\section{Additional Experiments}
\subsection{Ablation Study on TED Gesture Dataset}
We further conduct ablation study on TED Gesture and report results below, which shows the effectiveness of each module. The TED Gesture dataset lacks finger annotations, resulting in the lower motion hierarchy and less significant performance improvement brought by each module.

\begin{table}[h] \footnotesize
\setlength{\tabcolsep}{2.0pt}
\label{tbl:dy}
    \centering
        \begin{tabular}{c|c|c|c|c|c}
            \hline
             metric$\backslash$setting& $\bm{f}^\mathrm{high}_a$ only & Holistic & w/o $\mathcal{L}_{\mathrm{phy}}$ & HA2G-ASR & \textbf{HA2G Full} \\
            \hline
            FGD & 3.569 & 3.682 & 3.165 & 3.091 & \textbf{3.072} \\
            
            \hline
        \end{tabular}
        % \caption{Effect of AIB in different SNRs.}
        \vspace{-5.mm}
\end{table}

\subsection{Influence of Reference Frame Number $M$}
All the models are implemented with the same amount of information given as the input, including the number of initial poses $M$. The setting of using $M=4$ frames as seed pose is proposed in Trimodal~\cite{yoon2020speech}. Our whole setting basically follows theirs. To investigate the influence of $M$, we further set $M$ as 1 and 7. The results below suggest that the performance gain derived from additional initial poses is marginal, which shows the robustness of the proposed method to hyper-parameter $M$.

\setlength{\tabcolsep}{4pt}
\begin{table}[H]
    \centering
  \begin{tabular}{lccc}
    \toprule
    $M$ & FGD $\downarrow$ & BC $\uparrow$ & Diversity $\uparrow$ \\
    \midrule
     1 & 5.994 & 0.708 & 169.425\\
     4 & 5.306 & 0.715 & 173.899\\
     7 & 5.177 & 0.715 & 174.313\\
    \bottomrule
  \end{tabular}
  \caption{Influence of reference frame number $M$.}
    \label{framenumber}
\end{table}

\subsection{Randomness of Diversity Metric}
The Diversity metric is adapted from~\cite{NEURIPS2019_7ca57a9f} and is popularly used in other works~\cite{huang2020dance}. In order to mitigate the influence of randomness, we randomly sample 500 pairs, which is much more than the number 200 in~\cite{NEURIPS2019_7ca57a9f}. To ensure and verify the reproducibility, we further conduct the evaluation 10 times (create random samples 10 times with different random seeds). The results are listed in the table below. We can see that the difference is comparatively small between each group, which proves that the Diversity metric can be reproduced and the sample number of 500 is enough to alleviate randomness.

\setlength{\tabcolsep}{4pt}
\begin{table}[H]
    \centering
  \begin{tabular}{lccccc}
    \toprule
    Group & 1 & 2 & 3 & 4 & 5 \\
    \midrule
     Diversity & 172.58 & 171.91 & 173.60 & 173.66 & 173.71\\
     \midrule
     Group & 6 & 7 & 8 & 9 & 10 \\
     \midrule
     Diversity & 172.12 & 171.88 & 173.02 & 173.80 & 172.83\\
    \bottomrule
  \end{tabular}
  \caption{Randomness of the Diversity metric.}
    \label{randomness}
\end{table}

\section{Limitations and Future Work}
\begin{figure}
    \includegraphics[width=0.49\linewidth]{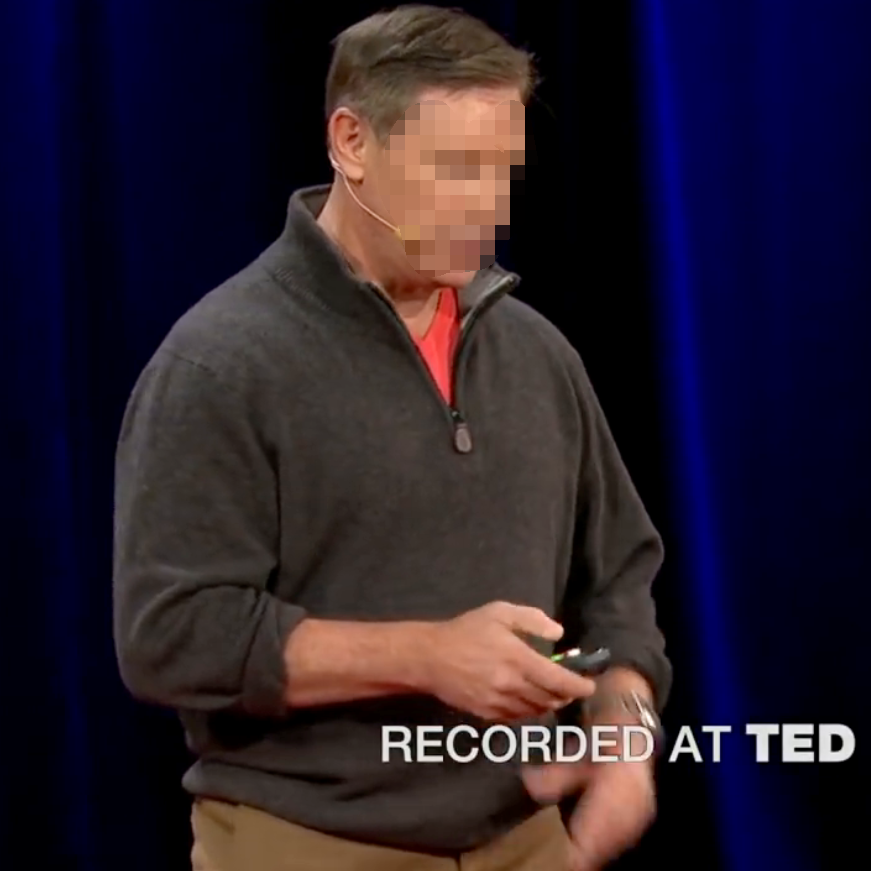}\hspace{-0.5mm}
    \includegraphics[width=0.49\linewidth]{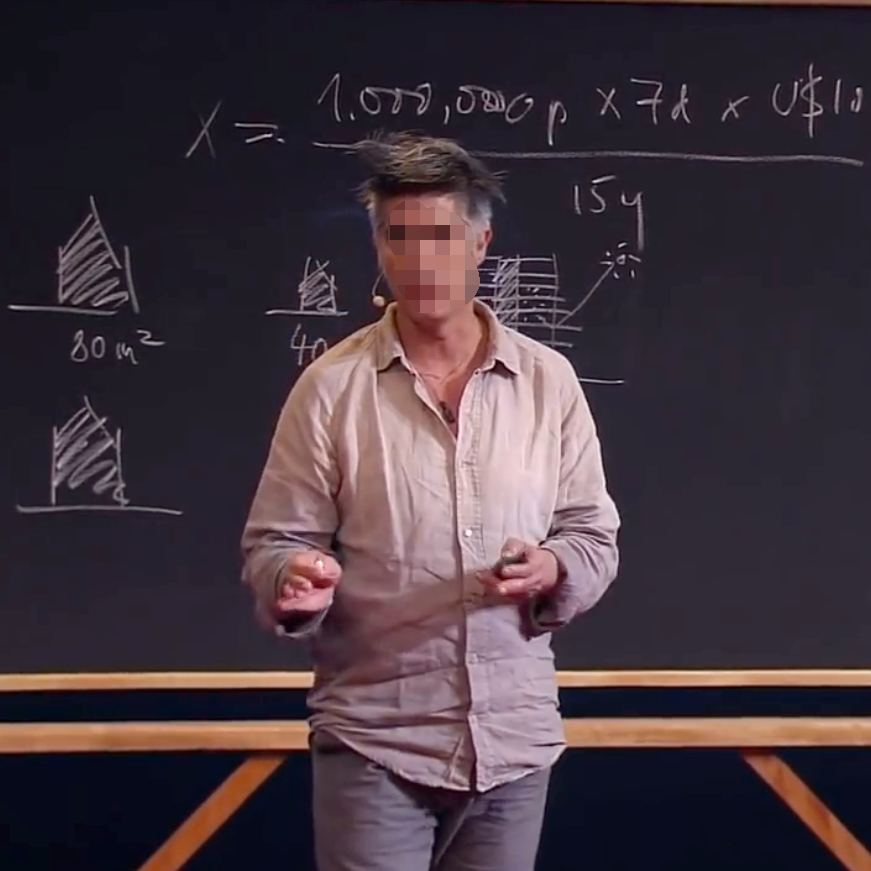}\hspace{-0.5mm}
    \caption{\textbf{Examples of data samples at specific setting with very strong prior on hand pose.} We implement the mosaic operations for all the images to eliminate personally identifiable information.}
    \label{limitation}
\end{figure}

Our work mainly have the following limitations: \textbf{1)} Since when people talk to others, the most important non-verbal behavior is upper body movements. Hence we only delve into the co-speech gesture generation of human upper body, without considering full body motions. This will make our trained avatars fail to walk around like TED Talk narrators. \textbf{2)} In the TED Talk dataset, some data samples have very strong prior on human hand pose at the specific settings that will affect training, \textit{e.g.}, people with speaker or chalk in their hand as shown in Fig.~\ref{limitation}. \textbf{3)} Although our proposed approach can capture the fine-grained motions of co-speech finger movements and diverse dynamic patterns of different human body parts, we still find it difficult to capture some very subtle movements like ``shrug''. This is mainly due to the fact that there hardly exists such action samples in the dataset and it is very hard for our model to learn such dynamic patterns. In future work, we will improve our method to capture full-body co-speech gestures and some very minor pose movements and we will enhance the automatic dataset pipeline algorithm to filter samples with strong prior that may affect our training quality.

\section{Social Impact}
Making co-speech gestures to complement conversational information is a kind of innate non-verbal behavior for human, while this work encourages the machine intelligence to be equipped with such ability, especially learn to animate the subtle hand and arm motions. Therefore, this work can exert positive impacts on both machine learning research and application field. On the one hand, the proposed approach identifies the advantages of hierarchical architecture design to extract cross-modal information at multiple granularities and excavate the fine-grained audio-pose associations, which can further facilitate cross-modal animation tasks like talking face generation and music2dance prospectively. On the other hand, the speech-driven gesture generation technique has a wide range of beneficial applications for society, including digital human broadcaster and social robots. Specifically, it could also assist dumb people to learn communication skills by teaching sign language with expressive human-like motions.
Since the generated motions are all skeleton-based, they hardly have detrimental impact in most cases. Still, it may potentially lead to the misuse of copyrighted 3D character models if we animate them without permission. Besides, the bias of the dataset may have some negative impact, \textit{e.g.}, some gestures may have negative meanings for some nations. But we believe the proper use of this technique will enhance positive societal development.

\section{Details of User Study}
The study involves 24 participants. They take 25-35 min to complete the task. The participants are 12 females and 12 males, with age range of 18-24 years old. The users are unaware of which motion sequence corresponds to which method or even the ground truth. Specifically, we randomly shuffle the order of video placement for all methods every time, so that participants can concentrate only on the quality of generated results for fair comparison. 

We have provided the users with instructions before conducting the study. The participants are asked to judge the three perspectives in the following manner: a) For “Naturalness”, does motion look natural and like real human poses regardless of background speech? There should not be any strange angles and unnatural movements. b) For “Smoothness”, does the generated motion maintain smoothness in temporal dimension, with no obvious rigid or stuck movements, regardless of background speech audio? c) For "Synchrony", does the generated motion match the corresponding speech audio both rhythmically and semantically? We also show the raw videos of TED Talk before participants’ rating process to help them make more accurate judgement. 24 participants of 12 females and 12 males are involved in the study, covering 4 nationalities in order to bridge biases. 12 of them are researchers from the field of deep generative models and others are from other fields.

\newpage

\end{document}